\documentclass[10pt,letterpaper]{article}

\usepackage{cogsci}

\cogscifinalcopy 

\usepackage{times}
\usepackage{latexsym}
\usepackage{amsmath}
\usepackage{algorithm}
\usepackage{algorithmic}
\usepackage{pslatex}
\usepackage{apacite}
\usepackage{array}
\usepackage{graphicx}
\usepackage{multirow}
\usepackage{appendix}
\usepackage{multicol} 
\usepackage{xcolor}
\definecolor{forestgreen}{rgb}{0, 0.5, 0}
\definecolor{yellow}{rgb}{1, 0, 0}
\definecolor{decision-color}{rgb}{0.92, 0.5, 0.98}
\definecolor{detection-color}{rgb}{0.25, 0.87, 0.80}
\definecolor{gbi-color}{rgb}{0.5, 0.9, 0.1}
\definecolor{sbi-color}{rgb}{1, 0.6, 0.07}
\usepackage{makecell}
\usepackage{float} 

\title{Balancing Rigor and Utility: Mitigating Cognitive Biases in Large Language Models for Multiple-Choice Questions}
 
\author{
Hanyang Zhong\thanks{These authors contributed equally to this work and share first authorship.}$^{1}$, 
Liman Wang\footnotemark[1]$^{1}$, 
Wenting Cao$^{2}$, 
Zeyuan Sun$^{3}$ \\
$^{1}$University of York \\
$^{2}$Friedrich Schiller University Jena \\
$^{3}$King's College London \\
\texttt{\{hanyang.zhong,liman.wang\}@york.ac.uk, wenting.cao@uni-jena.de, zeyuan.sun@kcl.ac.uk}
}

\begin{document}

\maketitle
\begin{abstract}
This paper examines the role of cognitive biases in the decision-making processes of large language models (LLMs), challenging the conventional goal of eliminating all biases. When properly balanced, we show that certain cognitive biases can enhance decision-making efficiency through rational deviations and heuristic shortcuts. By introducing heuristic moderation and an abstention option, which allows LLMs to withhold responses when uncertain, we reduce error rates, improve decision accuracy, and optimize decision rates. Using the Balance Rigor and Utility (BRU) dataset, developed through expert collaboration, our findings demonstrate that targeted inspection of cognitive biases aligns LLM decisions more closely with human reasoning, enhancing reliability and suggesting strategies for future improvements. This approach offers a novel way to leverage cognitive biases to improve the practical utility of LLMs across various applications.
\end{abstract}

\section{Introduction}

Bias in LLMs is a critical challenge in AI research. While significant efforts have been made to address social biases embedded in training datasets \cite{bang2024measuring, gallegos2024bias, minaee2024large}, cognitive biases that emerge during inference processes remain underexplored and problematic \cite{suri2024large, wang2024real, moore2024reasoningbiasstudycounterfactual}. These biases can mirror human cognitive tendencies, leading to flawed decision-making. Rational deviations, a concept from psychology introduced by Gerd Gigerenzer, suggest that not all biases are inherently harmful. These deviations involve heuristic thinking that simplifies decision-making, especially under uncertainty, but they can also introduce systematic errors \cite{Gigerenzer2006, Kruis2020, Berthet2022}. In LLMs, these deviations manifest in design choices that balance performance and efficiency. For example, models like GPT-4 \cite{openai2024gpt4} use context-driven token prediction to generate text, prioritizing coherence but sometimes at the cost of accuracy \cite{brown2020language}. These reasoning shortcuts mimic human cognitive biases, leading to suboptimal outcomes, making their mitigation essential for developing reliable and fair AI systems.

\begin{figure}[t]
  \includegraphics[width=\columnwidth]{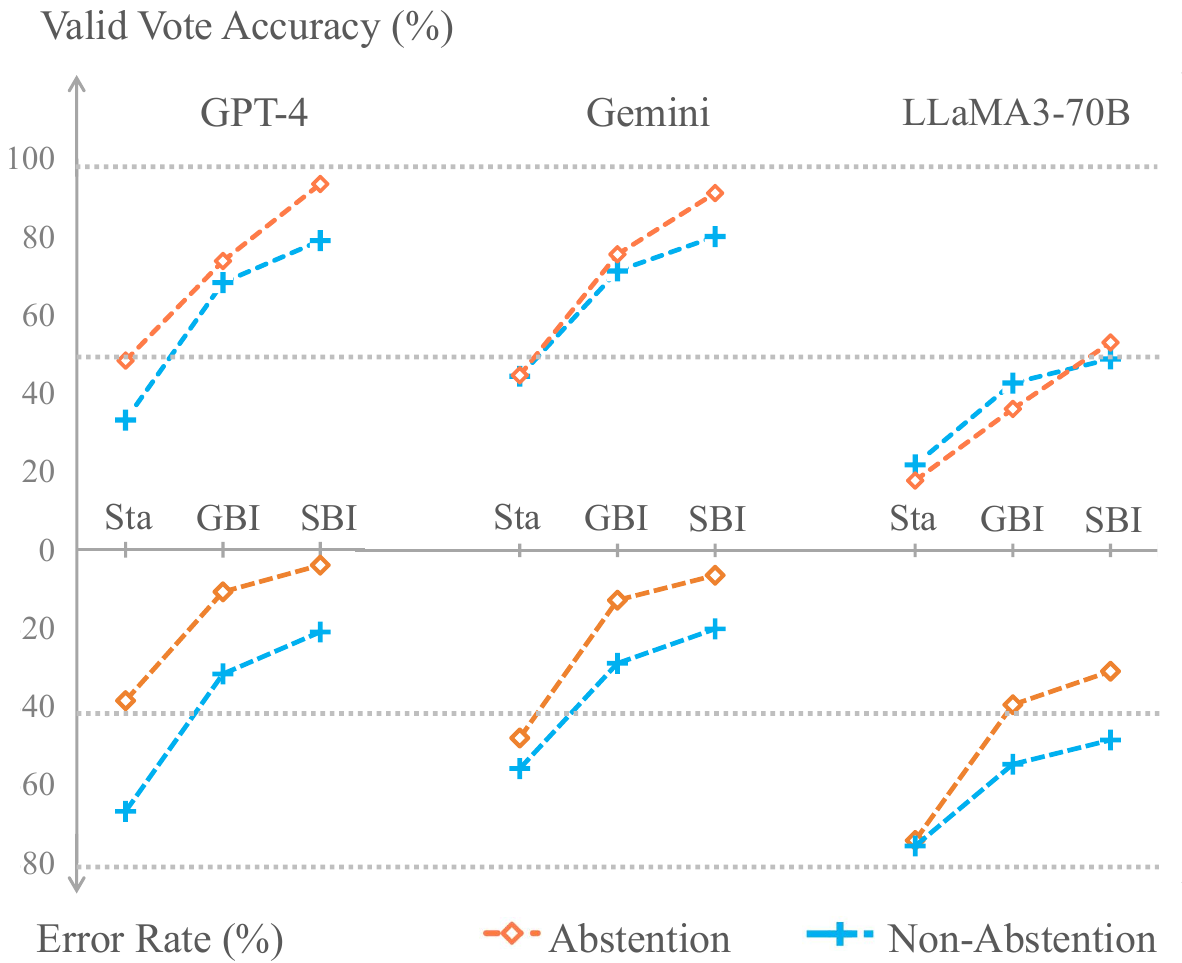}
  \caption{Valid vote accuracy and error rates on the BRU dataset for LLMs balancing rational deviations, both with and without the option to abstain. 'Sta' represents the standard baseline used for comparison, while 'GBI' and 'SBI' denote the proposed prompting strategies, as detailed in Section 4.}
  \label{fig:fig1}
\end{figure}

Our research tackles these challenges by proposing a balanced approach that integrates abstention as a rational response alongside cognitive bias mitigation techniques. We introduce heuristic moderation and an abstention mechanism, enabling LLMs to withhold decisions when uncertainty is high, reducing errors and improving accuracy. Using the BRU dataset, we demonstrate that scaling bias inspection and incorporating abstention significantly improve model performance and align LLM decision-making with human reasoning. Our result and dataset is available at: \url{hanyangzhong.github.io/BRU-website}.

\begin{figure*}[t]
\includegraphics[width=\linewidth]{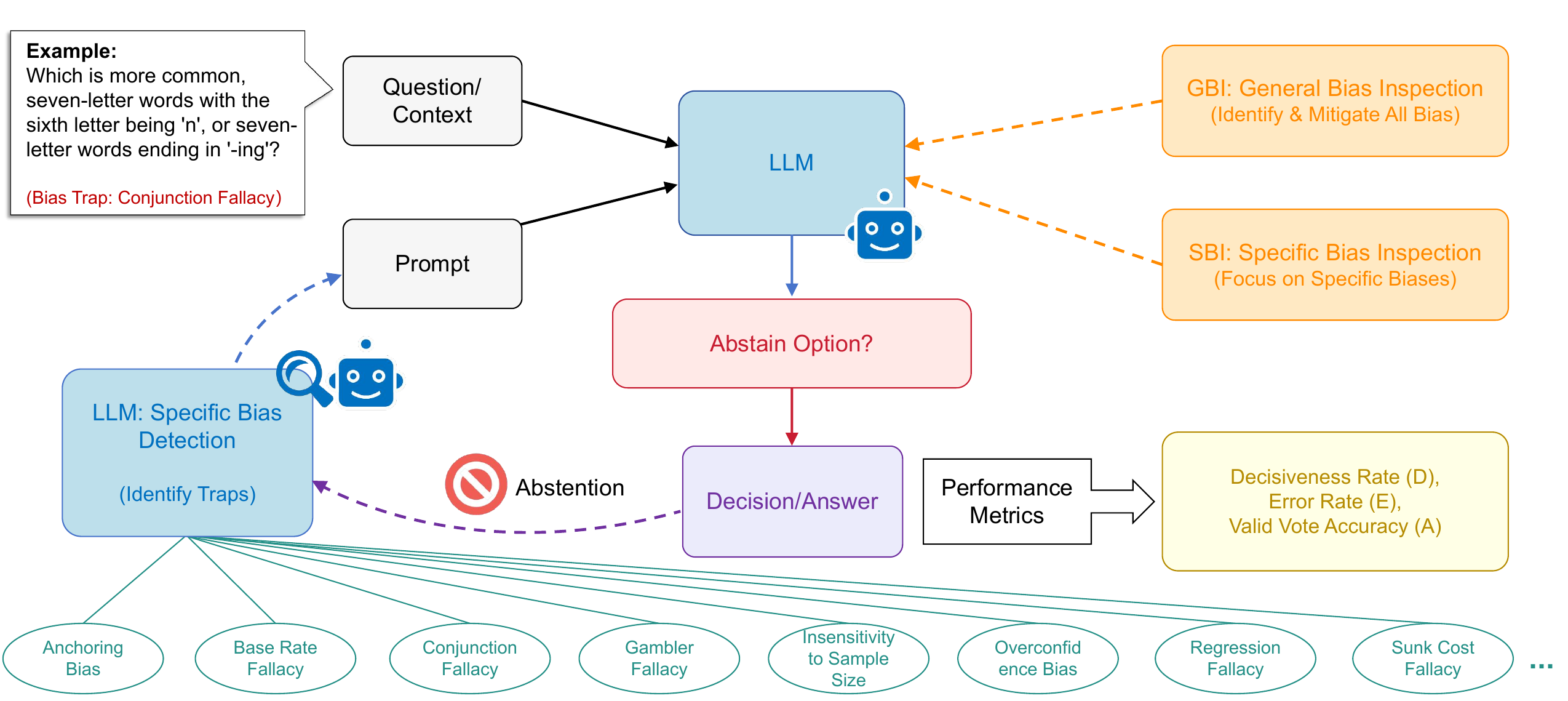}
\caption{QA examples from GPT-4. The Conjunction Fallacy is a subset of cognitive biases. \textbf{Scaling the scope of bias inspection can influence rational deviations,} thereby impacting the outcomes of LLMs' reasoning. To address this, we propose a feedback loop \textbf{Bias Detection} module to identify the type of bias and adjust the inspection scope when an abstention from answering is considered. This approach ensures that LLMs provide more accurate responses by systematically addressing biases during decision-making. The detailed demonstration of the whole workflow is shown in Appendix Table 18-21. }
\label{fig:fig2}
\end{figure*}

\section{Related Works}

Research on mitigating cognitive biases and rational deviations in LLMs is extensive. Suri et al. \cite{suri2024large} identify human-like biases in GPT-3.5, such as anchoring and framing effects, while Bubeck et al. \cite{Bubeck2023} and Binz and Schulz \cite{01Binz2023} observe similar issues, including the framing effect and conjunction fallacy, in GPT-4 and GPT-3. Wang et al. \cite{wang2024real} show that LLMs often rely on stereotypes over statistical reasoning, reflecting the representativeness heuristic.
Advancements suggest multi-modal and multi-agent systems \cite{jiang2024multimodal} as promising solutions for enhancing LLM rationality. Grounding decisions in broader sensory contexts, as demonstrated by Awadalla et al. \cite{Awadalla2023} and Bai et al. \cite{Bai2023}, helps reduce hallucinations and improve decision-making. Yang et al. \cite{Yang2024} and Wu et al. \cite{Wu2022} highlight external knowledge sources to address model limitations, while Du et al. \cite{Du2023} and Cohen et al. \cite{Cohen2023} emphasize consensus and debate mechanisms in multi-agent systems. Efforts to improve LLM reliability also stress robust evaluation metrics, as discussed by Echterhoff et al. \cite{Echterhoff2024} and Wang et al. \cite{Wang2024}. Wang et al. \cite{wang2024answersreviewingrationalitymultiple} further critique multiple-choice question answering (MCQA) benchmarks and propose the MCQA+ dataset for better performance assessment.
Building on this foundation, our research bridges the gap between rational deviations and cognitive biases by integrating heuristic moderation with abstention strategies, improving decision accuracy and robustness in LLM reasoning and prediction.

\section{Methodology}


\subsection{Phenomenon and Method Formation}

Evaluating LLMs through multiple-choice questions (MCQs) poses a key challenge: traditional metrics force definitive choices, even under uncertainty, ignoring the nuanced decision-making influenced by cognitive biases. These biases can aid efficiency via heuristics but also cause systematic errors \cite{wang2024real}.
Accuracy metrics often fail to differentiate between confident correct answers and correct guesses, obscuring a model’s true capabilities and skewing performance evaluations. For instance, high accuracy may stem from guessing rather than informed decisions, raising concerns about reliability in real-world use \cite{wang2024answersreviewingrationalitymultiple}.
To address this, we propose new metrics that better capture LLM decision-making complexities:

\begin{itemize}
    \item \textbf{Decisiveness Rate ($D$):} Captures the model's willingness to commit to an answer, providing insight into how often it prefers to abstain rather than make an uninformed guess.
    \item \textbf{Error Rate ($E$):} Unlike traditional metrics that simply count correct answers, this metric focuses on the frequency of incorrect choices, particularly under conditions of uncertainty.
    \item \textbf{Valid Vote Accuracy ($A$):} Measures the accuracy of the model's decisions when it chooses to answer, thereby highlighting its reliability in situations where it has confidence.
\end{itemize}

These new metrics aim to offer a more accurate and nuanced evaluation of LLM performance, balancing reducing error rates with improving decision accuracy. By allowing models to abstain when uncertain, we can better align LLM decision-making with human-like reasoning, ultimately enhancing their reliability and trustworthiness in practical applications \cite{madhusudhan2024llmsknowanswerinvestigating}.

\subsection{Strategic Abstention}

Abstaining is crucial for mitigating cognitive biases in LLMs during MCQs, enhancing accuracy and fairness. Forcing answers under uncertainty can lead to biased and erroneous responses, especially with skewed training data. Enabling abstention helps models avoid decisions that exacerbate biases, improving reliability \cite{madhusudhan2024llmsknowanswerinvestigating}.
This is particularly important in complex reasoning tasks, where biases are more influential. Traditional metrics rewarding confident answers without considering uncertainty can reinforce these biases. Abstention mechanisms allow models to withhold responses when detecting bias or low confidence \cite{bala, chen2023adaptationselfevaluationimproveselective}.
In critical fields like healthcare and finance, abstention reduces risks from biased or harmful answers, enhancing reliability. By focusing on confident responses, LLMs improve task performance. Techniques like Strict Prompting and Chain-of-Thought (CoT) effectively support abstention, balancing rigor and utility \cite{madhusudhan2024llmsknowanswerinvestigating, wei2023chainofthoughtpromptingelicitsreasoning}.
Our experiments show that incorporating abstention improves LLM decision-making in uncertain or ambiguous scenarios, ensuring fairness and trust, especially in high-stakes environments.

\subsection{Scaling the Inspection Scope}

Research shows that providing LLMs with cues about cognitive biases can improve accuracy in biased MCQs \cite{Echterhoff2024}. However, improper scaling of corrective behavior hinders balanced application of rational deviations. Correcting biases is akin to adjusting focus: precision is key, as overcorrection or under-correction reduces effectiveness.
Our findings indicate that LLMs become more cautious when defining broad concepts like "Cognitive Bias," often opting to "abstain," which slightly boosts accuracy but reduces decisiveness. In contrast, defining specific concepts like "Conjunction Fallacy" makes the model less cautious and more decisive. As shown in Fig.~\ref{fig:fig2}, scaling the self-inspection scope appropriately ensures a balanced approach to managing cognitive biases, improving both accuracy and decisiveness.

\subsection{Feedback Loop with Bias Detection}

As shown in Fig.~\ref{fig:fig2}, the "Bias Detection Module" is an additional component designed to enhance the decision-making accuracy of LLMs by identifying cognitive bias traps embedded in MCQs. This module leverages the advanced reasoning capabilities of GPT-4o to detect biases that could influence the model's responses. The detection process is outlined in Algorithm~\ref{alg:feedback_loop}.

\begin{algorithm}
\caption{Feedback Loops with Bias Detection}
\begin{algorithmic}[1]
\fontsize{9}{14}\selectfont
\STATE \textbf{Initialize:} $Bias \gets \textit{None}$, $LoopCount \gets 0$
\STATE $Decision \gets \textsc{AnswerModel}(\textit{MCQ})$

\WHILE{$LoopCount < \textit{MaxLoops}$ \textbf{and} $Decision = \textit{Abstain}$}
    \STATE $Bias \gets \textsc{DetectModel}(\textit{MCQ})$
    \STATE $Decision \gets \textsc{AnswerModel}(\textit{MCQ},\ $Bias$)$
    \STATE $LoopCount \gets LoopCount + 1$
\ENDWHILE

\STATE \textbf{Output:} $Decision$
\end{algorithmic}
\label{alg:feedback_loop}
\end{algorithm}


The process starts with GPT-4o detecting potential biases in the question and prioritizing the subtype most likely to impact decision-making. This requires a robust model like GPT-4o to identify subtle biases effectively.
A feedback loop iteratively refines decision-making by integrating bias detection into responses. The loop continues until a bias is found or a maximum number of iterations is reached, enabling dynamic adjustment and uncovering less obvious biases.

A limitation is that "abstention" is never a correct ground truth answer. If the $\textsc{AnswerModel}$ selects "abstention," the loop may continue until the iteration limit, delaying a valid answer. To address this, we restrict the loop to one iteration, allowing "abstention" as an option while refining the output without indefinite looping. Future work could explore multi-iteration loops and nuanced modifications to improve adaptive bias detection.

\section{Dataset and Experimental Setup}


\subsection{Dataset Setup}

The BRU dataset used in this study includes 205 MCQs, designed to comprehensively test cognitive biases in language models. Unlike datasets like MMLU \cite{mmlu}, which cover broad question categories but have limited relevance to cognitive biases, or TruthfulQA \cite{truthfulqa} and PIQA \cite{piqa}, which focus on factual correctness and commonsense reasoning, the BRU dataset addresses a wider range of cognitive biases.

Developed by an experienced psychologist with input from a medical data expert for reliability, the dataset was optimized by NLP specialists for clarity in testing LLMs. 
Each bias subcategory is well-documented with detailed descriptions, as shown in Table 1 and Appendix A (Tables 9-12, Figs. 1 and 5), ensuring transparency and traceability. 
These resources provide comprehensive insights into the dataset's composition and question design.

\begin{table}[]
\centering
\resizebox{\columnwidth}{!}{%
\begin{tabular}{ll}
\Xhline{1px}
{\color[HTML]{000000} \textbf{Misjudgment of Probability}} & {\color[HTML]{000000} \textbf{Errors in Judgment}} \\ \hline
Base Rate Fallacy (40)            & Regression Fallacy (35)  \\
Conjunction Fallacy (15)          & Anchoring Bias (20)      \\
Insensitivity to Sample Size (30) & Overconfidence Bias (30) \\
Gambler's Fallacy (20)            & Sunk Cost Fallacy (15)    \\  \Xhline{1px}
\end{tabular}%
}
\caption{Categories and Quantities in the BRU Dataset}
\label{tab:table1}
\end{table}

\subsection{Models and Prompting}
In our study, we evaluate the performance of three LLMs: GPT-4 \cite{openai2024gpt4}, Gemini 1.0 Pro \cite{Gemini}, and LLaMA3-70B \cite{llama3}. To thoroughly assess their capabilities and decision-making processes, we employ various prompting techniques to mitigate or encourage heuristic thinking. These techniques are crucial for understanding how LLMs navigate cognitive tasks and biases.
\paragraph{Abstention Prompting}
Abstention prompting allows the model to refrain from making a decision when uncertain:
\begin{quote}
\textit{If you prefer not to make a decisive choice, then select option E.} \\
\textit{E: I am not sure which choice is the best to select.}
\end{quote}
This reduces incorrect answers by avoiding guesses and improving accuracy by encouraging the model to make decisions only when confident. The option E is designed to aid the final result statistics.
\paragraph{Non-Abstention Prompting}
This prompting forces the model to make a choice:
\begin{quote}
\textit{You can only choose one option.}
\end{quote}
This evaluates the LLM's performance under pressure by requiring it to make decisions even when uncertain. It assesses the model's ability to handle forced-choice scenarios, revealing its strategies for making informed guesses despite uncertainty.

\paragraph{General Bias Inspection}

Consider the set of all cognitive biases, denoted as \( B = \{b_1, b_2, b_3, \ldots, b_n\} \), where each element \( b_i \) represents a distinct cognitive bias. General Bias Inspection (GBI) involves a comprehensive review of the entire set \( B \) to identify and mitigate any cognitive bias in decision-making. This approach ensures that the model is broadly aware of the full spectrum of cognitive biases, enabling it to self-check for potential bias influences. By reflecting on the general concept of cognitive bias, defined as any systematic deviation from rational judgment, GBI promotes overall bias mitigation by considering the union of all possible biases \( \cup B \):
\begin{quote}
\textit{Please provide a definition of cognitive bias and identify any instances of these biases in the decision-making process.}
\end{quote}

\paragraph{Specific Bias Inspection}

Specific Bias Inspection (SBI) focuses on a particular subset of cognitive biases, denoted as \( S \subseteq B \), where \( S \) represents a specific group of related biases relevant to the current context or question. For example, if \( S = \{b_3\} \) represents biases related to probability misjudgments, such as the Base Rate Fallacy and the Gambler’s Fallacy, SBI involves a focused analysis on this subset \( S \). This targeted approach allows the model to concentrate on the most pertinent biases for a given scenario, enhancing accuracy by mitigating the influence of specific biases \( \cup S \) rather than the entire set \( B \):
\begin{quote}
\textit{Please provide a definition of the Base Rate Fallacy, then identify any instances of this specific bias in the decision-making process. }
\end{quote}
This targeted approach enables more precise and context-sensitive bias analysis. 

\paragraph{Bias Detection Module}
To further enhance the model's ability to detect potential "bias traps" in questions, we introduce a dedicated prompt for the Bias Detection Module: 
\begin{quote}
\textit{Please identify which cognitive bias trap is contained in this question and return the cognitive bias type. The most likely cognitive bias trap is \_ .}
\end{quote}
This prompt enables the model to identify specific cognitive biases embedded in each problem. Although the model may list all relevant bias subtypes, each question is designed to target a particular bias subtype, creating a hierarchy of relevance. In our experiments, we select the highest-priority bias subtype as the output, ensuring that the most significant cognitive bias is accurately identified and addressed.

\begin{table*}[]
\centering
\setlength{\tabcolsep}{3.1mm}{
\begin{tabular}{ccccccc}
\Xhline{1px}
\multirow{2}{*}{Model / Accuracy $A$} & \multicolumn{3}{c}{+ Non-Abstention}                                                           & \multicolumn{3}{c}{+ Abstention}                                          \\ \cline{2-7} 
                       & Standard                      & GBI                  & SBI                           & Standard             & GBI                  & SBI                  \\ \hline
GPT-4                  & 33.2                          & 68.3 {\scriptsize \textcolor{forestgreen}{(+35.1)}}          & 79.0 {\scriptsize \textcolor{forestgreen}{(\textbf{+45.8})}}                     & 48.4                & 73.8 {\scriptsize \textcolor{forestgreen}{(+25.4)}}          & \textbf{93.5} {\scriptsize \textcolor{forestgreen}{(+45.1)}}          \\
Gemini 1.0 Pro                 & 44.4                          & 71.2 {\scriptsize \textcolor{forestgreen}{(+26.8)}}          & \textbf{80.0} {\scriptsize \textcolor{forestgreen}{(+35.6)}}                    & 44.6                & 75.5 {\scriptsize \textcolor{forestgreen}{(+30.9)}}          & 91.1 {\scriptsize \textcolor{forestgreen}{(\textbf{+46.5})}}          \\
LLaMA3-70B             & 22.4                          & 43.9 {\scriptsize \textcolor{forestgreen}{(+21.5)}}         & 50.2 {\scriptsize \textcolor{forestgreen}{(+27.8)}}                   & 18.3                 & 37.1 {\scriptsize \textcolor{forestgreen}{(+18.8)}}          & 54.8 {\scriptsize \textcolor{forestgreen}{(+36.5)}}      \\ \Xhline{1px}
\multicolumn{1}{l}{}   & \multicolumn{1}{l}{\textbf{}} & \multicolumn{1}{l}{} & \multicolumn{1}{l}{\textbf{}} & \multicolumn{1}{l}{} & \multicolumn{1}{l}{} & \multicolumn{1}{l}{} \\ \Xhline{1px}
\multirow{2}{*}{Model / Error Rate $E$} & \multicolumn{3}{c}{+ Non-Abstention}                                                    & \multicolumn{3}{c}{+ Abstention}                                   \\ \cline{2-7} 
                       & Standard                      & GBI                  & SBI                           & Standard             & GBI                  & SBI                  \\ \hline
GPT-4                  & 66.8                          & 31.7 {\scriptsize \textcolor{yellow}{(-35.1)}}        & 21.0  {\scriptsize \textcolor{yellow}{(\textbf{-45.8})}}                     & 38.5                 & 10.7 {\scriptsize \textcolor{yellow}{(-27.8)}}         & \multicolumn{1}{r}{\textbf{3.9} {\scriptsize \textcolor{yellow}{(-34.6)}}}            \\
Gemini 1.0 Pro                 & 55.6                          & 28.8 {\scriptsize \textcolor{yellow}{(-26.8)}}          & \textbf{20.0}  {\scriptsize \textcolor{yellow}{(-35.6)}}                     & 47.8                 & 12.7 {\scriptsize \textcolor{yellow}{(-35.1)}}          & \multicolumn{1}{r}{6.3 {\scriptsize \textcolor{yellow}{(-41.5)}}}           \\
LLaMA3-70B             & 77.6                          & 56.1 {\scriptsize \textcolor{yellow}{(-21.5)}}          & 49.8 {\scriptsize \textcolor{yellow}{(-27.8)}}                   & 76.1                 & 40.5 {\scriptsize \textcolor{yellow}{(-35.6)}}          & 31.7 {\scriptsize \textcolor{yellow}{(\textbf{-44.4})}}          \\ \Xhline{1px}
\end{tabular}}
\caption{Prediction accuracy and error rate of GPT-4, Gemini 1.0 Pro, and LLaMA3-70B in Non-Abstention and Abstention experiments (\%) on the BRU dataset with different prompting strategies. Bold numbers indicate the relative extrema. Differences between Standard groups with and without abstention are shown with $\pm$ values in black.}
\label{tab:table2}
\end{table*}

\subsection{Evaluation Criteria}

\begin{figure*}[t]
  \includegraphics[width=1\linewidth]{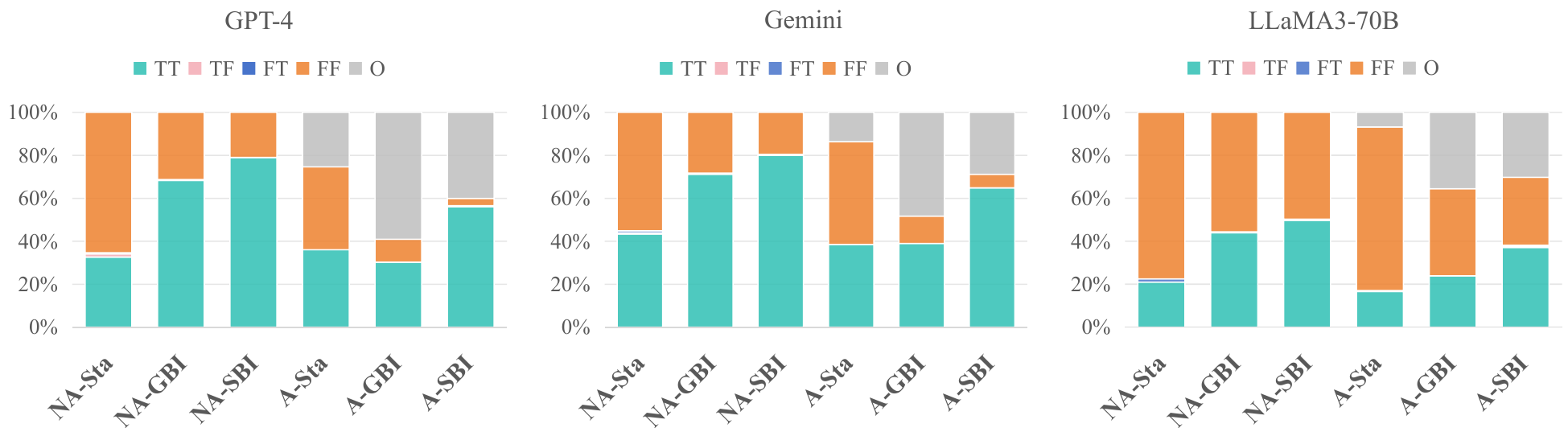}
  \caption{The combination of TT, TF, FT, FF, and O rates for GPT-4, Gemini 1.0 Pro, and LLaMA3-70B on the BRU dataset using different prompting strategies. 'NA-' denotes Non-Abstention, 'A-' denotes Abstention, and 'Sta' represents the Standard used for comparison. The detailed distributions of the TT, TF, FT, FF, and O rates for GPT-4, Gemini 1.0 Pro and LLaMA3-70B are elaborately listed in Appendix Tables 5, 6 and 7.} 
  \label{fig:fig4}
\end{figure*}

In evaluating the reasoning outcomes of LLMs, we categorize the reasoning process and results using the notations TT, TF, FT, and FF. These represent the following scenarios: a correct reasoning process with a correct result (TT), a correct reasoning process with an incorrect result (TF), an incorrect reasoning process with a correct result (FT), and an incorrect reasoning process with an incorrect result (FF). The symbol O denotes instances of "abstention," where the model selects option E.
For the statistical analysis, the Decisiveness Rate $D$ is defined as follows:

{\small \begin{align*}
D=\frac{N_{total}-N_{O}}{N_{total} } 
\end{align*}
\label{equate1}
}
where $N_{O}$ represents the number of abstained questions and $N_{total}$ represents the total number of questions in the BRU dataset. The Error Rate $E$ is defined as:

{\small \begin{align*}
E = \frac{N_{FF}+N_{TF} }{N_{total}-N_{O}  } 
\end{align*}
}

Here, $N_{FF}$ and $N_{TF}$ denote the number of questions with incorrect results. The Valid Vote Accuracy $A$ is given by:

{\small \begin{align*}
A=\frac{N_{TT}+N_{FT}}{N_{total}-N_{O} } 
\end{align*}
}

where $N_{TT}$ and $N_{FT}$ indicate the number of questions with correct results. It should be noted that the ground truth answers are manually annotated, excluding the reasoning process content. The accuracy of the LLM's reasoning is assessed through a manual review of the dialogue context.

\section{Experimental Results}

\subsection{Scaling Bias Inspection Effect Findings}
In the context of Non-Abstention, with "Standard" serving as the control group, significant improvements in $A$ score are observed for GPT-4, Gemini 1.0 Pro, and LLaMA3-70B on the BRU dataset when employing the prompting techniques GBI and SBI, as shown in Table.~\ref{tab:table2}. Notably, the $A$ score enhancement is more pronounced with SBI. Specifically, GPT-4, Gemini 1.0 Pro, and LLaMA3-70B achieve $A$ score of 79\%, 80\%, and 50.2\%, respectively, when utilizing SBI in Non-Abstention. This underscores the effectiveness of the GBI and SBI prompting strategies. The heightened accuracy with SBI, attributed to its narrower and more targeted cognitive bias scope, supports our hypothesis that specific guidance in bias inspection can significantly boost the accuracy of LLM responses.

\begin{figure*}[t]
  \includegraphics[width=1\linewidth]{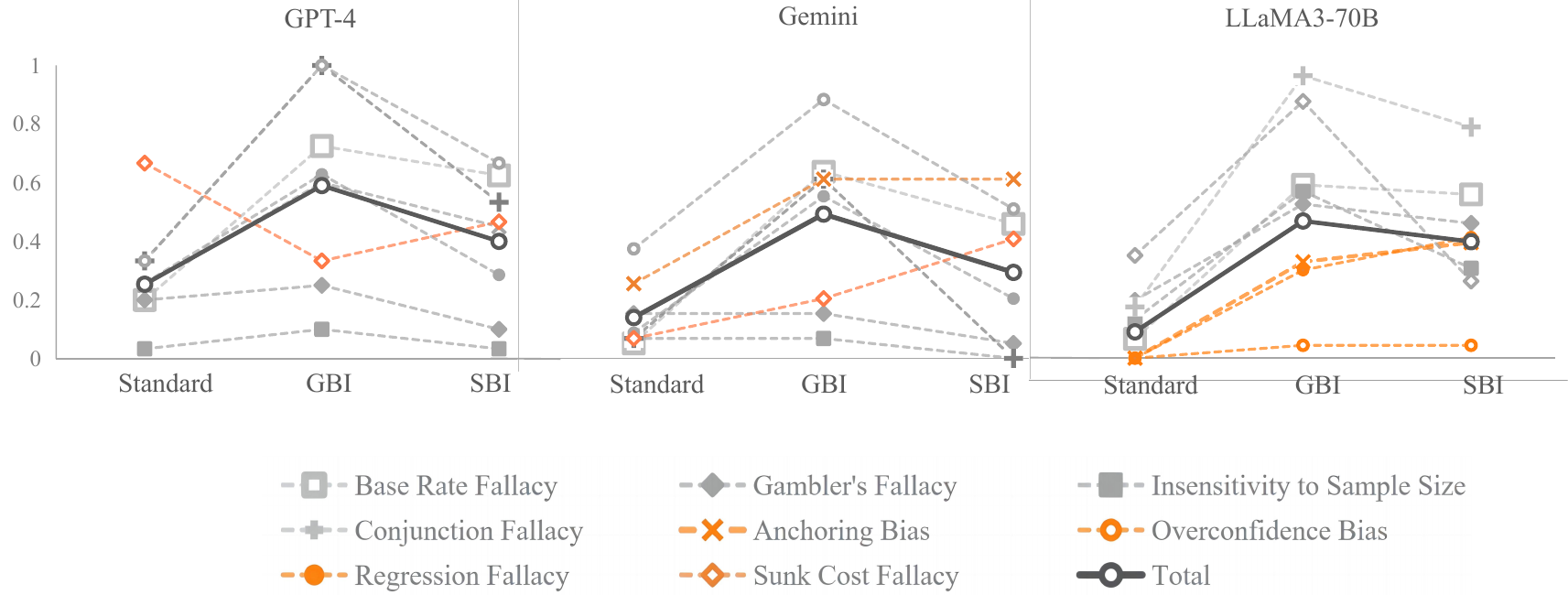}
  \caption{Distribution chart of abstention rates for GPT-4, Gemini 1.0 Pro, and LLaMA3-70B across different question types in the BRU dataset with Abstention enabled and using different prompting strategies.}
  \label{fig:fig5}
\end{figure*}

\subsection{Abstention Effect Findings}
As shown in Table~\ref{tab:table2}, the abstention effect reveals significant shifts in model performance. (see Appendix Tables 3, 4). 
Introducing the abstention option generally boosts the $A$ score but reduces the decision number, as depicted by the grey bars in Fig.~\ref{fig:fig4}.
For GPT-4, the $A$ score improves from 33.2\% to 48.4\%, indicating effective avoidance of incorrect answers through abstention. Gemini's $A$ score also sees a slight increase from 44.4\% to 44.6\%. Conversely, LLaMA3-70B's $A$ score drops from 22.4\% to 18.3\%, highlighting its weaker decision-making capabilities.
$E$ score decreases notably with abstention: GPT-4's drops from 66.8\% to 38.5\%, and Gemini 1.0 Pro's from 55.6\% to 47.8\%. LLaMA3-70B experiences a minor reduction from 77.6\% to 76.1\%. These results suggest that abstention reduces errors and leads to fewer decisions.

\subsection{Combination Test Findings}
Before conducting the combination tests, we assessed whether the Abstention technique offered greater benefits than GBI and SBI. As shown in Table \ref{tab:table2}, the improvements from abstention (highlighted in green) are generally less significant than those from GBI and SBI.
However, GPT-4 gains more from abstention rights than Gemini 1.0 Pro and LLaMA3-70B. Under the Abstention condition, GPT-4 performs best with SBI, while Gemini 1.0 Pro peaks with SBI under the Non-Abstention condition. Fig.~\ref{fig:fig4} also shows that GPT-4 abstains more frequently than Gemini 1.0 Pro, adopting a "fewer decisions, fewer mistakes" strategy. Although this approach indicates emergent intelligence, excessive abstention has potential risks, which will be discussed further.
In Fig.~\ref{fig:fig4}, the green bars represent the proportion of correct answers, which increases with GBI and SBI but decreases with Abstention, consistent with earlier findings. Conversely, the $E$ decreases when Abstention, GBI, or SBI are introduced.
Among the tested combinations, \textbf{Abstention+SBI} allows all three LLMs to achieve the highest $A$ score of \textbf{93.5\%} and the lowest $E$ score of \textbf{3.9\%}, as shown in Table \ref{tab:table2}. Notably, GPT-4 and Gemini 1.0 Pro using Abstention+SBI achieve near-zero $E$ scores of \textbf{3.9\%} and \textbf{6.3\%}, respectively, demonstrating the value of minimizing decision risk.

\subsection{Bias Detection Loop by GPT-4o}

Using GPT-4o for bias detection in 205 questions, the model accurately identified the specific bias subtype 65\% of the time. In an additional 15\% of cases, it matched a parent category or synonym of the bias subtype (details in Appendix Table 8)
, resulting in a cumulative \textbf{80\%} recognition rate. This demonstrates the module's ability to align closely with actual biases, effectively bridging GBI and SBI methods.
This high recognition rate enables dynamic adjustment of the bias detection scope, enhancing decision accuracy and minimizing errors. By refining bias recognition, the model seamlessly transitions from broader GBI to precise SBI, improving $D$ scores and reducing $E$ scores.

\subsection{Abstention Patterns for Question Types}

When LLMs are allowed to abstain, their abstention rates generally fluctuate across question types, as shown in Fig.~\ref{fig:fig5}. However, question types like the Sunk Cost Fallacy, Anchoring Bias, Overconfidence Bias, and Regression Fallacy (highlighted in orange) deviate from this trend. These Errors in Judgment, less common in training data than Misjudgment of Probability concepts, are more likely to be overlooked.
Under GBI alone, LLMs may miss these traps, leading to overconfidence and lower abstention rates. With SBI, the model recognizes these biases, shifting from overconfidence to uncertainty, reflected in higher abstention rates. This highlights how identifying specific biases, such as the "Sunk Cost Fallacy," prompts LLMs to reconsider their judgments.
The Bias Detection Loop module addresses this by using GPT-4o's reasoning to transition GBI abstention into automatic SBI recognition, boosting both $D$ and $A$ scores.

\section{Conclusion}

This study examines the roles of cognitive biases and rational deviations in LLM decision-making, showing that leveraging certain biases through heuristic moderation and strategic abstention can be beneficial. The BRU dataset reveals that SBI prompting, combined with the option to abstain from uncertain decisions, reduces errors and aligns LLM reasoning with human patterns.
The findings demonstrate that integrating abstention with targeted bias inspection, like SBI, significantly enhances accuracy and reliability. Models such as GPT-4 and Gemini 1.0 Pro achieve notable improvements in decision accuracy by abstaining under uncertainty, better replicating human-like judgment. The proposed Bias Detection Loop facilitates a seamless transition from GBI to SBI, combining inspection scopes without over-relying on abstention, further improving accuracy and utility.
This research underscores the value of balanced bias management in LLMs, showing that biases, when properly moderated, can enhance practical utility in applications like conversational agents and decision support systems. It establishes a foundation for further exploration of how balancing biases and rational deviations can improve LLM performance, particularly in MCQs.


\bibliographystyle{apacite}

\setlength{\bibleftmargin}{.125in}
\setlength{\bibindent}{-\bibleftmargin}

\bibliography{CogSci_Template}

\begin{thebibliography}{}

\bibitem [\protect \citeauthoryear {%
Arkes%
\ \BBA {} Blumer%
}{%
Arkes%
\ \BBA {} Blumer%
}{%
{\protect \APACyear {1985}}%
}]{%
Sunkcost}
\APACinsertmetastar {%
Sunkcost}%
\begin{APACrefauthors}%
Arkes, H\BPBI R.%
\BCBT {}\ \BBA {} Blumer, C.%
\end{APACrefauthors}%
\unskip\
\newblock
\APACrefYearMonthDay{1985}{}{}.
\newblock
{\BBOQ}\APACrefatitle {The psychology of sunk cost} {The psychology of sunk cost}.{\BBCQ}
\newblock
\APACjournalVolNumPages{Organizational Behavior and Human Decision Processes}{35}{1}{124-140}.
\PrintBackRefs{\CurrentBib}

\bibitem [\protect \citeauthoryear {%
Awadalla%
\ \protect \BOthers {.}}{%
Awadalla%
\ \protect \BOthers {.}}{%
{\protect \APACyear {2023}}%
}]{%
Awadalla2023}
\APACinsertmetastar {%
Awadalla2023}%
\begin{APACrefauthors}%
Awadalla, A.%
, Gao, I.%
, Gardner, J.%
, Hessel, J.%
, Hanafy, Y.%
, Zhu, W.%
\BDBL {}Schmidt, L.%
\end{APACrefauthors}%
\unskip\
\newblock
\APACrefYearMonthDay{2023}{}{}.
\newblock
\APACrefbtitle {OpenFlamingo: An Open-Source Framework for Training Large Autoregressive Vision-Language Models.} {Openflamingo: An open-source framework for training large autoregressive vision-language models.}
\newblock
\begin{APACrefURL} \url{https://arxiv.org/abs/2308.01390} \end{APACrefURL}
\PrintBackRefs{\CurrentBib}

\bibitem [\protect \citeauthoryear {%
Bai%
\ \protect \BOthers {.}}{%
Bai%
\ \protect \BOthers {.}}{%
{\protect \APACyear {2023}}%
}]{%
Bai2023}
\APACinsertmetastar {%
Bai2023}%
\begin{APACrefauthors}%
Bai, J.%
, Bai, S.%
, Yang, S.%
, Wang, S.%
, Tan, S.%
, Wang, P.%
\BDBL {}Zhou, J.%
\end{APACrefauthors}%
\unskip\
\newblock
\APACrefYearMonthDay{2023}{}{}.
\newblock
\APACrefbtitle {Qwen-VL: A Versatile Vision-Language Model for Understanding, Localization, Text Reading, and Beyond.} {Qwen-vl: A versatile vision-language model for understanding, localization, text reading, and beyond.}
\newblock
\begin{APACrefURL} \url{https://arxiv.org/abs/2308.12966} \end{APACrefURL}
\PrintBackRefs{\CurrentBib}

\bibitem [\protect \citeauthoryear {%
Balabanov%
\ \BBA {} Linander%
}{%
Balabanov%
\ \BBA {} Linander%
}{%
{\protect \APACyear {2024}}%
}]{%
bala}
\APACinsertmetastar {%
bala}%
\begin{APACrefauthors}%
Balabanov, O.%
\BCBT {}\ \BBA {} Linander, H.%
\end{APACrefauthors}%
\unskip\
\newblock
\APACrefYearMonthDay{2024}{}{}.
\newblock
\APACrefbtitle {Uncertainty quantification in fine-tuned LLMs using LoRA ensembles.} {Uncertainty quantification in fine-tuned llms using lora ensembles.}
\newblock
\begin{APACrefURL} \url{https://arxiv.org/abs/2402.12264} \end{APACrefURL}
\PrintBackRefs{\CurrentBib}

\bibitem [\protect \citeauthoryear {%
Bang%
, Chen%
, Lee%
\BCBL {}\ \BBA {} Fung%
}{%
Bang%
\ \protect \BOthers {.}}{%
{\protect \APACyear {2024}}%
}]{%
bang2024measuring}
\APACinsertmetastar {%
bang2024measuring}%
\begin{APACrefauthors}%
Bang, Y.%
, Chen, D.%
, Lee, N.%
\BCBL {}\ \BBA {} Fung, P.%
\end{APACrefauthors}%
\unskip\
\newblock
\APACrefYearMonthDay{2024}{}{}.
\newblock
{\BBOQ}\APACrefatitle {Measuring political bias in large language models: What is said and how it is said} {Measuring political bias in large language models: What is said and how it is said}.{\BBCQ}
\newblock
\BIn{} \APACrefbtitle {Proceedings of the 62nd {Annual} {Meeting} of the {Association} for {Computational} {Linguistics} ({Volume} 1: Long {Papers})} {Proceedings of the 62nd {Annual} {Meeting} of the {Association} for {Computational} {Linguistics} ({Volume} 1: Long {Papers})}\ (\BPGS\ 11142--11159).
\newblock
\APACaddressPublisher{Stroudsburg, PA, USA}{}.
\PrintBackRefs{\CurrentBib}

\bibitem [\protect \citeauthoryear {%
Berthet%
}{%
Berthet%
}{%
{\protect \APACyear {2022}}%
}]{%
Berthet2022}
\APACinsertmetastar {%
Berthet2022}%
\begin{APACrefauthors}%
Berthet, V.%
\end{APACrefauthors}%
\unskip\
\newblock
\APACrefYearMonthDay{2022}{}{}.
\newblock
{\BBOQ}\APACrefatitle {The Impact of Cognitive Biases on Professionals’ Decision-Making: A Review of Four Occupational Areas} {The impact of cognitive biases on professionals’ decision-making: A review of four occupational areas}.{\BBCQ}
\newblock
\APACjournalVolNumPages{Frontiers in Psychology}{12}{}{802439}.
\newblock
\begin{APACrefURL} \url{https://doi.org/10.3389/fpsyg.2021.802439} \end{APACrefURL}
\PrintBackRefs{\CurrentBib}

\bibitem [\protect \citeauthoryear {%
Binz%
\ \BBA {} Schulz%
}{%
Binz%
\ \BBA {} Schulz%
}{%
{\protect \APACyear {2023}}%
}]{%
01Binz2023}
\APACinsertmetastar {%
01Binz2023}%
\begin{APACrefauthors}%
Binz, M.%
\BCBT {}\ \BBA {} Schulz, E.%
\end{APACrefauthors}%
\unskip\
\newblock
\APACrefYearMonthDay{2023}{}{}.
\newblock
{\BBOQ}\APACrefatitle {Using Cognitive Psychology to Understand {GPT-3}} {Using cognitive psychology to understand {GPT-3}}.{\BBCQ}
\newblock
\APACjournalVolNumPages{Proceedings of the National Academy of Sciences}{120}{6}{e2218523120}.
\PrintBackRefs{\CurrentBib}

\bibitem [\protect \citeauthoryear {%
Bisk%
, Zellers%
, Bras%
, Gao%
\BCBL {}\ \BBA {} Choi%
}{%
Bisk%
\ \protect \BOthers {.}}{%
{\protect \APACyear {2020}}%
}]{%
piqa}
\APACinsertmetastar {%
piqa}%
\begin{APACrefauthors}%
Bisk, Y.%
, Zellers, R.%
, Bras, R\BPBI L.%
, Gao, J.%
\BCBL {}\ \BBA {} Choi, Y.%
\end{APACrefauthors}%
\unskip\
\newblock
\APACrefYearMonthDay{2020}{}{}.
\newblock
{\BBOQ}\APACrefatitle {{PIQA:} Reasoning about Physical Commonsense in Natural Language} {{PIQA:} reasoning about physical commonsense in natural language}.{\BBCQ}
\newblock
\APACaddressPublisher{}{{AAAI} Press}.
\newblock
\begin{APACrefURL} \url{https://doi.org/10.1609/aaai.v34i05.6239} \end{APACrefURL}
\PrintBackRefs{\CurrentBib}

\bibitem [\protect \citeauthoryear {%
Brown%
\ \protect \BOthers {.}}{%
Brown%
\ \protect \BOthers {.}}{%
{\protect \APACyear {2020}}%
}]{%
brown2020language}
\APACinsertmetastar {%
brown2020language}%
\begin{APACrefauthors}%
Brown, T\BPBI B.%
, Mann, B.%
, Ryder, N.%
, Subbiah, M.%
, Kaplan, J.%
, Dhariwal, P.%
\BDBL {}Amodei, D.%
\end{APACrefauthors}%
\unskip\
\newblock
\APACrefYearMonthDay{2020}{}{}.
\newblock
{\BBOQ}\APACrefatitle {Language models are few-shot learners} {Language models are few-shot learners}.{\BBCQ}
\newblock
\BIn{} \APACrefbtitle {Proceedings of the 34th International Conference on Neural Information Processing Systems.} {Proceedings of the 34th international conference on neural information processing systems.}
\newblock
\APACaddressPublisher{}{Curran Associates Inc.}
\PrintBackRefs{\CurrentBib}

\bibitem [\protect \citeauthoryear {%
Bubeck%
\ \protect \BOthers {.}}{%
Bubeck%
\ \protect \BOthers {.}}{%
{\protect \APACyear {2023}}%
}]{%
Bubeck2023}
\APACinsertmetastar {%
Bubeck2023}%
\begin{APACrefauthors}%
Bubeck, S.%
, Chandrasekaran, V.%
, Eldan, R.%
, Gehrke, J.%
, Horvitz, E.%
, Kamar, E.%
\BDBL {}Zhang, Y.%
\end{APACrefauthors}%
\unskip\
\newblock
\APACrefYearMonthDay{2023}{}{}.
\newblock
\APACrefbtitle {Sparks of Artificial General Intelligence: Early experiments with GPT-4.} {Sparks of artificial general intelligence: Early experiments with gpt-4.}
\newblock
\begin{APACrefURL} \url{https://arxiv.org/abs/2303.12712} \end{APACrefURL}
\PrintBackRefs{\CurrentBib}

\bibitem [\protect \citeauthoryear {%
Chen%
\ \protect \BOthers {.}}{%
Chen%
\ \protect \BOthers {.}}{%
{\protect \APACyear {2023}}%
}]{%
chen2023adaptationselfevaluationimproveselective}
\APACinsertmetastar {%
chen2023adaptationselfevaluationimproveselective}%
\begin{APACrefauthors}%
Chen, J.%
, Yoon, J.%
, Ebrahimi, S.%
, Arik, S.%
, Pfister, T.%
\BCBL {}\ \BBA {} Jha, S.%
\end{APACrefauthors}%
\unskip\
\newblock
\APACrefYearMonthDay{2023}{}{}.
\newblock
{\BBOQ}\APACrefatitle {Adaptation with {Self}-{Evaluation} to {Improve} {Selective} {Prediction} in {LLMs}} {Adaptation with {Self}-{Evaluation} to {Improve} {Selective} {Prediction} in {LLMs}}.{\BBCQ}
\newblock
\BIn{} \APACrefbtitle {Findings of the {Association} for {Computational} {Linguistics}: EMNLP 2023.} {Findings of the {Association} for {Computational} {Linguistics}: Emnlp 2023.}
\newblock
\APACaddressPublisher{Stroudsburg, PA, USA}{}.
\PrintBackRefs{\CurrentBib}

\bibitem [\protect \citeauthoryear {%
Cohen%
, Hamri%
, Geva%
\BCBL {}\ \BBA {} Globerson%
}{%
Cohen%
\ \protect \BOthers {.}}{%
{\protect \APACyear {2023}}%
}]{%
Cohen2023}
\APACinsertmetastar {%
Cohen2023}%
\begin{APACrefauthors}%
Cohen, R.%
, Hamri, M.%
, Geva, M.%
\BCBL {}\ \BBA {} Globerson, A.%
\end{APACrefauthors}%
\unskip\
\newblock
\APACrefYearMonthDay{2023}{}{}.
\newblock
{\BBOQ}\APACrefatitle {{LM} vs {LM}: Detecting Factual Errors via Cross Examination} {{LM} vs {LM}: Detecting factual errors via cross examination}.{\BBCQ}
\newblock
\BIn{} \APACrefbtitle {Proceedings of the 2023 Conference on Empirical Methods in Natural Language Processing} {Proceedings of the 2023 conference on empirical methods in natural language processing}\ (\BPGS\ 12621--12640).
\newblock
\APACaddressPublisher{Singapore}{Association for Computational Linguistics}.
\newblock
\begin{APACrefURL} \url{https://aclanthology.org/2023.emnlp-main.778/} \end{APACrefURL}
\PrintBackRefs{\CurrentBib}

\bibitem [\protect \citeauthoryear {%
Du%
, Li%
, Torralba%
, Tenenbaum%
\BCBL {}\ \BBA {} Mordatch%
}{%
Du%
\ \protect \BOthers {.}}{%
{\protect \APACyear {2024}}%
}]{%
Du2023}
\APACinsertmetastar {%
Du2023}%
\begin{APACrefauthors}%
Du, Y.%
, Li, S.%
, Torralba, A.%
, Tenenbaum, J\BPBI B.%
\BCBL {}\ \BBA {} Mordatch, I.%
\end{APACrefauthors}%
\unskip\
\newblock
\APACrefYearMonthDay{2024}{}{}.
\newblock
{\BBOQ}\APACrefatitle {Improving factuality and reasoning in language models through multiagent debate} {Improving factuality and reasoning in language models through multiagent debate}.{\BBCQ}
\newblock
\BIn{} \APACrefbtitle {Proceedings of the 41st International Conference on Machine Learning.} {Proceedings of the 41st international conference on machine learning.}
\newblock
\APACaddressPublisher{}{JMLR.org}.
\PrintBackRefs{\CurrentBib}

\bibitem [\protect \citeauthoryear {%
Echterhoff%
, Liu%
, Alessa%
, McAuley%
\BCBL {}\ \BBA {} He%
}{%
Echterhoff%
\ \protect \BOthers {.}}{%
{\protect \APACyear {2024}}%
}]{%
Echterhoff2024}
\APACinsertmetastar {%
Echterhoff2024}%
\begin{APACrefauthors}%
Echterhoff, J\BPBI M.%
, Liu, Y.%
, Alessa, A.%
, McAuley, J.%
\BCBL {}\ \BBA {} He, Z.%
\end{APACrefauthors}%
\unskip\
\newblock
\APACrefYearMonthDay{2024}{}{}.
\newblock
{\BBOQ}\APACrefatitle {Cognitive Bias in Decision-Making with {LLM}s} {Cognitive bias in decision-making with {LLM}s}.{\BBCQ}
\newblock
\BIn{} (\BPGS\ 12640--12653).
\newblock
\APACaddressPublisher{}{Association for Computational Linguistics}.
\newblock
\begin{APACrefURL} \url{https://aclanthology.org/2024.findings-emnlp.739/} \end{APACrefURL}
\PrintBackRefs{\CurrentBib}

\bibitem [\protect \citeauthoryear {%
Gallegos%
\ \protect \BOthers {.}}{%
Gallegos%
\ \protect \BOthers {.}}{%
{\protect \APACyear {2024}}%
}]{%
gallegos2024bias}
\APACinsertmetastar {%
gallegos2024bias}%
\begin{APACrefauthors}%
Gallegos, I\BPBI O.%
, Rossi, R\BPBI A.%
, Barrow, J.%
, Tanjim, M\BPBI M.%
, Kim, S.%
, Dernoncourt, F.%
\BDBL {}Ahmed, N\BPBI K.%
\end{APACrefauthors}%
\unskip\
\newblock
\APACrefYearMonthDay{2024}{}{}.
\newblock
{\BBOQ}\APACrefatitle {Bias and Fairness in Large Language Models: A Survey} {Bias and fairness in large language models: A survey}.{\BBCQ}
\newblock
\APACjournalVolNumPages{Computational Linguistics}{50}{3}{1097--1179}.
\newblock
\begin{APACrefURL} \url{https://aclanthology.org/2024.cl-3.8/} \end{APACrefURL}
\PrintBackRefs{\CurrentBib}

\bibitem [\protect \citeauthoryear {%
Gerd%
}{%
Gerd%
}{%
{\protect \APACyear {2006}}%
}]{%
Gigerenzer2006}
\APACinsertmetastar {%
Gigerenzer2006}%
\begin{APACrefauthors}%
Gerd, G.%
\end{APACrefauthors}%
\unskip\
\newblock
\APACrefYearMonthDay{2006}{}{}.
\newblock
{\BBOQ}\APACrefatitle {Bounded and Rational} {Bounded and rational}.{\BBCQ}
\newblock
\BIn{} R.~Stainton\ (\BED), \APACrefbtitle {Contemporary Debates in Cognitive Science} {Contemporary debates in cognitive science}\ (\BPGS\ 115--133).
\newblock
\APACaddressPublisher{}{Wiley-Blackwell}.
\PrintBackRefs{\CurrentBib}

\bibitem [\protect \citeauthoryear {%
Gilovich%
, Vallone%
\BCBL {}\ \BBA {} Tversky%
}{%
Gilovich%
\ \protect \BOthers {.}}{%
{\protect \APACyear {1985}}%
}]{%
Regress}
\APACinsertmetastar {%
Regress}%
\begin{APACrefauthors}%
Gilovich, T.%
, Vallone, R.%
\BCBL {}\ \BBA {} Tversky, A.%
\end{APACrefauthors}%
\unskip\
\newblock
\APACrefYearMonthDay{1985}{}{}.
\newblock
{\BBOQ}\APACrefatitle {The hot hand in basketball: On the misperception of random sequences} {The hot hand in basketball: On the misperception of random sequences}.{\BBCQ}
\newblock
\APACjournalVolNumPages{Cognitive Psychology}{17}{3}{295-314}.
\PrintBackRefs{\CurrentBib}

\bibitem [\protect \citeauthoryear {%
Hendrycks%
\ \protect \BOthers {.}}{%
Hendrycks%
\ \protect \BOthers {.}}{%
{\protect \APACyear {2021}}%
}]{%
mmlu}
\APACinsertmetastar {%
mmlu}%
\begin{APACrefauthors}%
Hendrycks, D.%
, Burns, C.%
, Basart, S.%
, Zou, A.%
, Mazeika, M.%
, Song, D.%
\BCBL {}\ \BBA {} Steinhardt, J.%
\end{APACrefauthors}%
\unskip\
\newblock
\APACrefYearMonthDay{2021}{}{}.
\newblock
{\BBOQ}\APACrefatitle {Measuring Massive Multitask Language Understanding} {Measuring massive multitask language understanding}.{\BBCQ}
\newblock
\BIn{} \APACrefbtitle {International Conference on Learning Representations.} {International conference on learning representations.}
\PrintBackRefs{\CurrentBib}

\bibitem [\protect \citeauthoryear {%
Jiang%
\ \protect \BOthers {.}}{%
Jiang%
\ \protect \BOthers {.}}{%
{\protect \APACyear {2024}}%
}]{%
jiang2024multimodal}
\APACinsertmetastar {%
jiang2024multimodal}%
\begin{APACrefauthors}%
Jiang, B.%
, Xie, Y.%
, Wang, X.%
, Su, W\BPBI J.%
, Taylor, C\BPBI J.%
\BCBL {}\ \BBA {} Mallick, T.%
\end{APACrefauthors}%
\unskip\
\newblock
\APACrefYearMonthDay{2024}{}{}.
\newblock
{\BBOQ}\APACrefatitle {Multi-Modal and Multi-Agent Systems Meet Rationality: A Survey} {Multi-modal and multi-agent systems meet rationality: A survey}.{\BBCQ}
\newblock
\BIn{} \APACrefbtitle {ICML 2024 Workshop on LLMs and Cognition.} {Icml 2024 workshop on llms and cognition.}
\PrintBackRefs{\CurrentBib}

\bibitem [\protect \citeauthoryear {%
Kahneman%
}{%
Kahneman%
}{%
{\protect \APACyear {2011}}%
}]{%
kahneman2011thinking}
\APACinsertmetastar {%
kahneman2011thinking}%
\begin{APACrefauthors}%
Kahneman, D.%
\end{APACrefauthors}%
\unskip\
\newblock
\APACrefYear{2011}.
\newblock
\APACrefbtitle {Thinking, fast and slow} {Thinking, fast and slow}.
\newblock
\APACaddressPublisher{}{macmillan}.
\PrintBackRefs{\CurrentBib}

\bibitem [\protect \citeauthoryear {%
Kruis%
, Maris%
, Marsman%
, Bolsinova%
\BCBL {}\ \BBA {} Maas%
}{%
Kruis%
\ \protect \BOthers {.}}{%
{\protect \APACyear {2020}}%
}]{%
Kruis2020}
\APACinsertmetastar {%
Kruis2020}%
\begin{APACrefauthors}%
Kruis, J.%
, Maris, G.%
, Marsman, M.%
, Bolsinova, M.%
\BCBL {}\ \BBA {} Maas, H.%
\end{APACrefauthors}%
\unskip\
\newblock
\APACrefYearMonthDay{2020}{}{}.
\newblock
{\BBOQ}\APACrefatitle {Deviations of rational choice: an integrative explanation of the endowment and several context effects} {Deviations of rational choice: an integrative explanation of the endowment and several context effects}.{\BBCQ}
\newblock
\APACjournalVolNumPages{Scientific Reports}{10}{}{}.
\PrintBackRefs{\CurrentBib}

\bibitem [\protect \citeauthoryear {%
Lin%
, Hilton%
\BCBL {}\ \BBA {} Evans%
}{%
Lin%
\ \protect \BOthers {.}}{%
{\protect \APACyear {2022}}%
}]{%
truthfulqa}
\APACinsertmetastar {%
truthfulqa}%
\begin{APACrefauthors}%
Lin, S.%
, Hilton, J.%
\BCBL {}\ \BBA {} Evans, O.%
\end{APACrefauthors}%
\unskip\
\newblock
\APACrefYearMonthDay{2022}{}{}.
\newblock
{\BBOQ}\APACrefatitle {{T}ruthful{QA}: Measuring How Models Mimic Human Falsehoods} {{T}ruthful{QA}: Measuring how models mimic human falsehoods}.{\BBCQ}
\newblock
\APACaddressPublisher{Dublin, Ireland}{Association for Computational Linguistics}.
\PrintBackRefs{\CurrentBib}

\bibitem [\protect \citeauthoryear {%
Madhusudhan%
, Madhusudhan%
, Yadav%
\BCBL {}\ \BBA {} Hashemi%
}{%
Madhusudhan%
\ \protect \BOthers {.}}{%
{\protect \APACyear {2024}}%
}]{%
madhusudhan2024llmsknowanswerinvestigating}
\APACinsertmetastar {%
madhusudhan2024llmsknowanswerinvestigating}%
\begin{APACrefauthors}%
Madhusudhan, N.%
, Madhusudhan, S\BPBI T.%
, Yadav, V.%
\BCBL {}\ \BBA {} Hashemi, M.%
\end{APACrefauthors}%
\unskip\
\newblock
\APACrefYearMonthDay{2024}{}{}.
\newblock
\APACrefbtitle {Do LLMs Know When to NOT Answer? Investigating Abstention Abilities of Large Language Models.} {Do llms know when to not answer? investigating abstention abilities of large language models.}
\newblock
\begin{APACrefURL} \url{https://arxiv.org/abs/2407.16221} \end{APACrefURL}
\PrintBackRefs{\CurrentBib}

\bibitem [\protect \citeauthoryear {%
Meta%
}{%
Meta%
}{%
{\protect \APACyear {2024}}%
}]{%
llama3}
\APACinsertmetastar {%
llama3}%
\begin{APACrefauthors}%
Meta.%
\end{APACrefauthors}%
\unskip\
\newblock
\APACrefYearMonthDay{2024}{}{}.
\newblock
\APACrefbtitle {Introducing {Meta} {Llama} 3: The most capable openly available {LLM} to date.} {Introducing {Meta} {Llama} 3: The most capable openly available {LLM} to date.}
\newblock
\APAChowpublished {https://ai.meta.com/blog/meta-llama-3}.
\PrintBackRefs{\CurrentBib}

\bibitem [\protect \citeauthoryear {%
Minaee%
\ \protect \BOthers {.}}{%
Minaee%
\ \protect \BOthers {.}}{%
{\protect \APACyear {2025}}%
}]{%
minaee2024large}
\APACinsertmetastar {%
minaee2024large}%
\begin{APACrefauthors}%
Minaee, S.%
, Mikolov, T.%
, Nikzad, N.%
, Chenaghlu, M.%
, Socher, R.%
, Amatriain, X.%
\BCBL {}\ \BBA {} Gao, J.%
\end{APACrefauthors}%
\unskip\
\newblock
\APACrefYearMonthDay{2025}{}{}.
\newblock
\APACrefbtitle {Large Language Models: A Survey.} {Large language models: A survey.}
\newblock
\begin{APACrefURL} \url{https://arxiv.org/abs/2402.06196} \end{APACrefURL}
\PrintBackRefs{\CurrentBib}

\bibitem [\protect \citeauthoryear {%
Moore%
, Roberts%
, Pham%
\BCBL {}\ \BBA {} Fisher%
}{%
Moore%
\ \protect \BOthers {.}}{%
{\protect \APACyear {2024}}%
}]{%
moore2024reasoningbiasstudycounterfactual}
\APACinsertmetastar {%
moore2024reasoningbiasstudycounterfactual}%
\begin{APACrefauthors}%
Moore, K.%
, Roberts, J.%
, Pham, T.%
\BCBL {}\ \BBA {} Fisher, D.%
\end{APACrefauthors}%
\unskip\
\newblock
\APACrefYearMonthDay{2024}{}{}.
\newblock
\APACrefbtitle {Reasoning Beyond Bias: A Study on Counterfactual Prompting and Chain of Thought Reasoning.} {Reasoning beyond bias: A study on counterfactual prompting and chain of thought reasoning.}
\newblock
\begin{APACrefURL} \url{https://arxiv.org/abs/2408.08651} \end{APACrefURL}
\PrintBackRefs{\CurrentBib}

\bibitem [\protect \citeauthoryear {%
OpenAI%
\ \protect \BOthers {.}}{%
OpenAI%
\ \protect \BOthers {.}}{%
{\protect \APACyear {2024}}%
}]{%
openai2024gpt4}
\APACinsertmetastar {%
openai2024gpt4}%
\begin{APACrefauthors}%
OpenAI%
, Achiam, J.%
, Adler, S.%
, Agarwal, S.%
, Ahmad, L.%
, Akkaya, I.%
\BDBL {}Zoph, B.%
\end{APACrefauthors}%
\unskip\
\newblock
\APACrefYearMonthDay{2024}{}{}.
\newblock
\APACrefbtitle {GPT-4 Technical Report.} {Gpt-4 technical report.}
\newblock
\begin{APACrefURL} \url{https://arxiv.org/abs/2303.08774} \end{APACrefURL}
\PrintBackRefs{\CurrentBib}

\bibitem [\protect \citeauthoryear {%
Pichai%
\ \BBA {} Hassabis%
}{%
Pichai%
\ \BBA {} Hassabis%
}{%
{\protect \APACyear {2023}}%
}]{%
Gemini}
\APACinsertmetastar {%
Gemini}%
\begin{APACrefauthors}%
Pichai, S.%
\BCBT {}\ \BBA {} Hassabis, D.%
\end{APACrefauthors}%
\unskip\
\newblock
\APACrefYearMonthDay{2023}{}{}.
\newblock
\APACrefbtitle {Introducing Gemini: our largest and most capable AI model.} {Introducing gemini: our largest and most capable ai model.}
\newblock
\begin{APACrefURL} \url{https://blog.google/technology/ai/google-gemini-ai} \end{APACrefURL}
\newblock
\APACrefnote{Accessed: April 9, 2025}
\PrintBackRefs{\CurrentBib}

\bibitem [\protect \citeauthoryear {%
Suri%
, Slater%
, Ziaee%
\BCBL {}\ \BBA {} Nguyen%
}{%
Suri%
\ \protect \BOthers {.}}{%
{\protect \APACyear {2024}}%
}]{%
suri2024large}
\APACinsertmetastar {%
suri2024large}%
\begin{APACrefauthors}%
Suri, G.%
, Slater, L.%
, Ziaee, A.%
\BCBL {}\ \BBA {} Nguyen, M.%
\end{APACrefauthors}%
\unskip\
\newblock
\APACrefYearMonthDay{2024}{}{}.
\newblock
{\BBOQ}\APACrefatitle {Do Large Language Models Show Decision Heuristics Similar to Humans? A Case Study Using GPT-3.5} {Do large language models show decision heuristics similar to humans? a case study using gpt-3.5}.{\BBCQ}
\newblock
\APACjournalVolNumPages{Journal of Experimental Psychology: General}{153}{}{1066-1075}.
\PrintBackRefs{\CurrentBib}

\bibitem [\protect \citeauthoryear {%
Tversky%
\ \BBA {} Kahneman%
}{%
Tversky%
\ \BBA {} Kahneman%
}{%
{\protect \APACyear {1974}}%
}]{%
tversky1974judgment}
\APACinsertmetastar {%
tversky1974judgment}%
\begin{APACrefauthors}%
Tversky, A.%
\BCBT {}\ \BBA {} Kahneman, D.%
\end{APACrefauthors}%
\unskip\
\newblock
\APACrefYearMonthDay{1974}{}{}.
\newblock
{\BBOQ}\APACrefatitle {Judgment under Uncertainty: Heuristics and Biases: Biases in Judgments Reveal Some Heuristics of Thinking under Uncertainty} {Judgment under uncertainty: Heuristics and biases: Biases in judgments reveal some heuristics of thinking under uncertainty}.{\BBCQ}
\newblock
\APACjournalVolNumPages{Science}{185}{4157}{1124--1131}.
\PrintBackRefs{\CurrentBib}

\bibitem [\protect \citeauthoryear {%
Tversky%
\ \BBA {} Kahneman%
}{%
Tversky%
\ \BBA {} Kahneman%
}{%
{\protect \APACyear {1983}}%
}]{%
Conj}
\APACinsertmetastar {%
Conj}%
\begin{APACrefauthors}%
Tversky, A.%
\BCBT {}\ \BBA {} Kahneman, D.%
\end{APACrefauthors}%
\unskip\
\newblock
\APACrefYearMonthDay{1983}{}{}.
\newblock
{\BBOQ}\APACrefatitle {Extensional versus intuitive reasoning: The conjunction fallacy in probability judgment} {Extensional versus intuitive reasoning: The conjunction fallacy in probability judgment}.{\BBCQ}
\newblock
\APACjournalVolNumPages{Psychological Review}{90}{4}{293--315}.
\newblock
\begin{APACrefDOI} \doi{10.1037/0033-295X.90.4.293} \end{APACrefDOI}
\PrintBackRefs{\CurrentBib}

\bibitem [\protect \citeauthoryear {%
H.~Wang%
, Zhao%
, Qiang%
, Qin%
\BCBL {}\ \BBA {} Liu%
}{%
H.~Wang%
\ \protect \BOthers {.}}{%
{\protect \APACyear {2024}}%
}]{%
wang2024answersreviewingrationalitymultiple}
\APACinsertmetastar {%
wang2024answersreviewingrationalitymultiple}%
\begin{APACrefauthors}%
Wang, H.%
, Zhao, S.%
, Qiang, Z.%
, Qin, B.%
\BCBL {}\ \BBA {} Liu, T.%
\end{APACrefauthors}%
\unskip\
\newblock
\APACrefYearMonthDay{2024}{}{}.
\newblock
{\BBOQ}\APACrefatitle {Beyond the Answers: Reviewing the Rationality of Multiple Choice Question Answering for the Evaluation of Large Language Models} {Beyond the answers: Reviewing the rationality of multiple choice question answering for the evaluation of large language models}.{\BBCQ}
\newblock
\APACjournalVolNumPages{CoRR}{abs/2402.01349}{}{}.
\newblock
\begin{APACrefURL} \url{https://doi.org/10.48550/arXiv.2402.01349} \end{APACrefURL}
\PrintBackRefs{\CurrentBib}

\bibitem [\protect \citeauthoryear {%
P.~Wang%
, Xiao%
, Chen%
\BCBL {}\ \BBA {} Oswald%
}{%
P.~Wang%
\ \protect \BOthers {.}}{%
{\protect \APACyear {2024}}%
}]{%
wang2024real}
\APACinsertmetastar {%
wang2024real}%
\begin{APACrefauthors}%
Wang, P.%
, Xiao, Z.%
, Chen, H.%
\BCBL {}\ \BBA {} Oswald, F\BPBI L.%
\end{APACrefauthors}%
\unskip\
\newblock
\APACrefYearMonthDay{2024}{}{}.
\newblock
{\BBOQ}\APACrefatitle {Will the Real Linda Please Stand up...to Large Language Models? Examining the Representativeness Heuristic in {LLM}s} {Will the real linda please stand up...to large language models? examining the representativeness heuristic in {LLM}s}.{\BBCQ}
\newblock
\BIn{} \APACrefbtitle {First Conference on Language Modeling.} {First conference on language modeling.}
\newblock
\begin{APACrefURL} \url{https://openreview.net/forum?id=3GhOWfSLrD} \end{APACrefURL}
\PrintBackRefs{\CurrentBib}

\bibitem [\protect \citeauthoryear {%
S.~Wang%
, Long%
, Fan%
, Huang%
\BCBL {}\ \BBA {} Wei%
}{%
S.~Wang%
\ \protect \BOthers {.}}{%
{\protect \APACyear {2025}}%
}]{%
Wang2024}
\APACinsertmetastar {%
Wang2024}%
\begin{APACrefauthors}%
Wang, S.%
, Long, Z.%
, Fan, Z.%
, Huang, X.%
\BCBL {}\ \BBA {} Wei, Z.%
\end{APACrefauthors}%
\unskip\
\newblock
\APACrefYearMonthDay{2025}{}{}.
\newblock
{\BBOQ}\APACrefatitle {Benchmark Self-Evolving: A Multi-Agent Framework for Dynamic {LLM} Evaluation} {Benchmark self-evolving: A multi-agent framework for dynamic {LLM} evaluation}.{\BBCQ}
\newblock
\BIn{} \APACrefbtitle {Proceedings of the 31st International Conference on Computational Linguistics} {Proceedings of the 31st international conference on computational linguistics}\ (\BPGS\ 3310--3328).
\newblock
\APACaddressPublisher{Abu Dhabi, UAE}{Association for Computational Linguistics}.
\newblock
\begin{APACrefURL} \url{https://aclanthology.org/2025.coling-main.223/} \end{APACrefURL}
\PrintBackRefs{\CurrentBib}

\bibitem [\protect \citeauthoryear {%
Wei%
\ \protect \BOthers {.}}{%
Wei%
\ \protect \BOthers {.}}{%
{\protect \APACyear {2022}}%
}]{%
wei2023chainofthoughtpromptingelicitsreasoning}
\APACinsertmetastar {%
wei2023chainofthoughtpromptingelicitsreasoning}%
\begin{APACrefauthors}%
Wei, J.%
, Wang, X.%
, Schuurmans, D.%
, Bosma, M.%
, Ichter, B.%
, Xia, F.%
\BDBL {}Zhou, D.%
\end{APACrefauthors}%
\unskip\
\newblock
\APACrefYearMonthDay{2022}{}{}.
\newblock
{\BBOQ}\APACrefatitle {Chain-of-thought prompting elicits reasoning in large language models} {Chain-of-thought prompting elicits reasoning in large language models}.{\BBCQ}
\newblock
\BIn{} \APACrefbtitle {Proceedings of the 36th International Conference on Neural Information Processing Systems.} {Proceedings of the 36th international conference on neural information processing systems.}
\newblock
\APACaddressPublisher{Red Hook, NY, USA}{Curran Associates Inc.}
\PrintBackRefs{\CurrentBib}

\bibitem [\protect \citeauthoryear {%
Wu%
, Lu%
, Sabharwal%
\BCBL {}\ \BBA {} Mottaghi%
}{%
Wu%
\ \protect \BOthers {.}}{%
{\protect \APACyear {2022}}%
}]{%
Wu2022}
\APACinsertmetastar {%
Wu2022}%
\begin{APACrefauthors}%
Wu, J.%
, Lu, J.%
, Sabharwal, A.%
\BCBL {}\ \BBA {} Mottaghi, R.%
\end{APACrefauthors}%
\unskip\
\newblock
\APACrefYearMonthDay{2022}{}{}.
\newblock
{\BBOQ}\APACrefatitle {Multi-Modal Answer Validation for Knowledge-Based VQA} {Multi-modal answer validation for knowledge-based vqa}.{\BBCQ}
\newblock
\APACjournalVolNumPages{Proceedings of the AAAI Conference on Artificial Intelligence}{}{}{}.
\PrintBackRefs{\CurrentBib}

\bibitem [\protect \citeauthoryear {%
Yang%
, Chen%
, Li%
, Wang%
\BCBL {}\ \BBA {} Yang%
}{%
Yang%
\ \protect \BOthers {.}}{%
{\protect \APACyear {2024}}%
}]{%
Yang2024}
\APACinsertmetastar {%
Yang2024}%
\begin{APACrefauthors}%
Yang, Z.%
, Chen, G.%
, Li, X.%
, Wang, W.%
\BCBL {}\ \BBA {} Yang, Y.%
\end{APACrefauthors}%
\unskip\
\newblock
\APACrefYearMonthDay{2024}{}{}.
\newblock
{\BBOQ}\APACrefatitle {DoraemonGPT: toward understanding dynamic scenes with large language models (exemplified as a video agent)} {Doraemongpt: toward understanding dynamic scenes with large language models (exemplified as a video agent)}.{\BBCQ}
\newblock
\BIn{} \APACrefbtitle {Proceedings of the 41st International Conference on Machine Learning.} {Proceedings of the 41st international conference on machine learning.}
\newblock
\APACaddressPublisher{}{JMLR.org}.
\PrintBackRefs{\CurrentBib}

\end{thebibliography}
\clearpage
\appendix

\section{Appendix}
\label{sec:appendix}

The appendix contains detailed information on the segmented performances of GPT-4, Gemini 1.0 Pro, and LLaMA3-70B in both the Non-Abstention and Abstention experiments. 

Please refer to Tables \ref{table3} and \ref{table4}, which expand upon the contents of Table \ref{tab:table2}.
Additionally, all experimental data for the Abstention experiments are included. This data covers correct and incorrect responses and abstention distributions across eight categories of bias. For further details, see Tables \ref{tab:table5}, \ref{tab:table6}, and \ref{tab:table7}, which supplement the information presented in Fig.~\ref{fig:fig4} of the main text.

In Tables \ref{tab:table5}, \ref{tab:table6}, and \ref{tab:table7}, bold type indicates the most prevalent response type within each category of bias, allowing for a clear visualization of how different models and prompts influence response distributions across the overall dataset.

\begin{figure*}[t]
\centering
  \includegraphics[width=\linewidth]{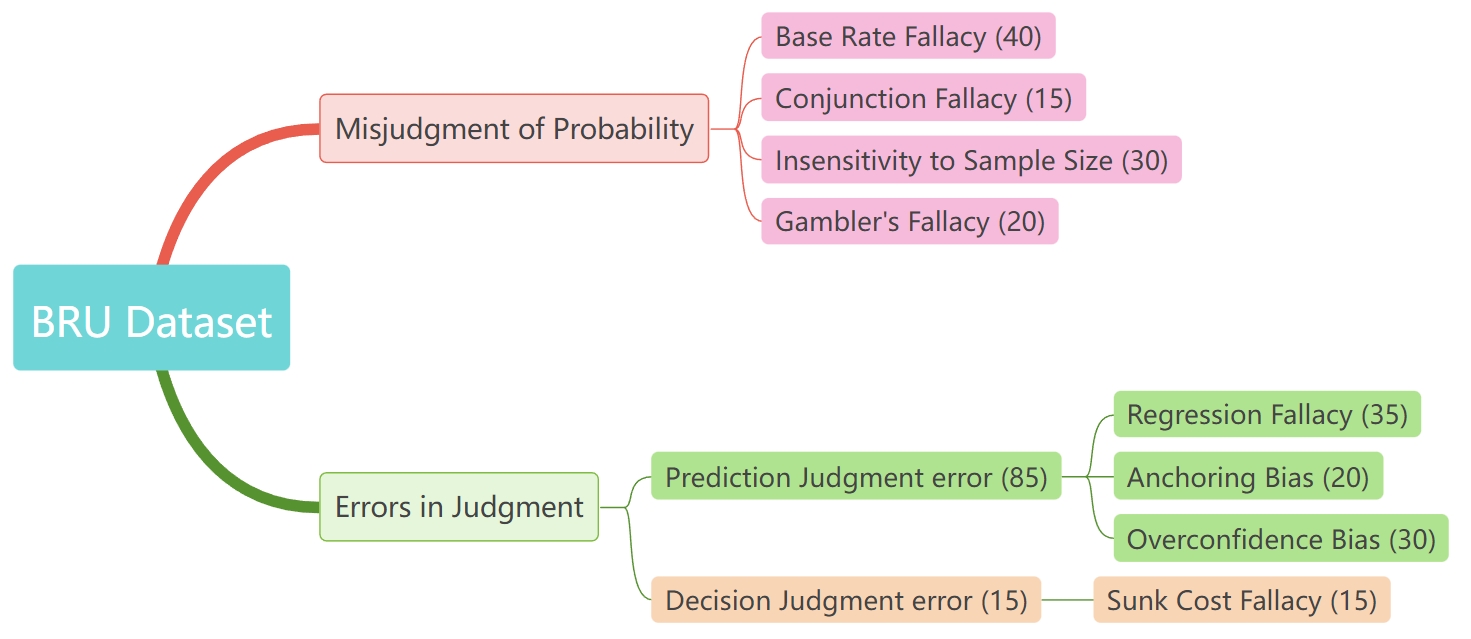}
  \caption{This diagram pertains to the specific details of dataset design and the classification of questions, with the numbers in parentheses indicating the quantity of questions in each category. }
  \label{fig:datasets}
\end{figure*}

Fig.~\ref{fig:datasets} illustrates the subcategories of the dataset. Tables \ref{table8}, \ref{table9}, \ref{table10}, and \ref{table11} provide detailed information about each type of bias in the dataset, further subdivisions, and corresponding question designs. The entire dataset’s question design primarily derives from these examples.

Tables \ref{table12}, \ref{table13}, and \ref{table14} display the performance of GPT-4 in the Abstention experiments, showing the effects of different prompts on the same questions and the correct answers to these questions. 

Tables \ref{table15} and \ref{table16} illustrate two different scenarios in the Bias Detection Module's question analysis: direct detection of the corresponding subtype and indirect detection of the subtype. Additionally, Tables \ref{table-demo1-1} through \ref{table-demo2-2} provide examples of the overall process as described in Fig.~\ref{fig:fig2}.

The complete dataset, including questions related to various biases, their corresponding Ground Truths, and detailed results of the experiments mentioned in this paper, will be published on GitHub following the paper's acceptance.

\begin{table*}[]
\centering
\setlength{\tabcolsep}{1.7mm}{
\begin{tabular}{lcccccc}
\Xhline{1.5px}
\multicolumn{7}{c}{Non-Abstention}                                                                                                                                                                                       \\ \Xhline{1.2px}
\multicolumn{1}{c}{\multirow{2}{*}{GPT-4}}      & \multicolumn{2}{c}{\begin{tabular}[c]{@{}c@{}}Standard\end{tabular}} & \multicolumn{2}{c}{GBI}                     & \multicolumn{2}{c}{SBI}                     \\ \cline{2-7} 
\multicolumn{1}{c}{}                            & Accuracy                             & Error Rate                          & Accuracy             & Error Rate           & Accuracy             & Error Rate           \\ \hline
Base Rate Fallacy                               & 17.5                                 & 82.5                                & 60.0                 & 40.0                 & 52.5                 & 47.5                 \\
Gambler's Fallacy                               & 75.0                                 & 25.0                                & 95.0                 & 5.0                  & 100.0                & 0.0                  \\
Insensitivity to Sample Size                    & 20.0                                 & 80.0                                & 73.3                 & 26.7                 & 96.7                 & 3.3                  \\
Conjunction Fallacy                             & 73.3                                 & 26.7                                & 93.3                 & 6.7                  & 80.0                 & 20.0                 \\
Anchoring Bias                                  & 80.0                                 & 20.0                                & 70.0                 & 30.0                 & 80.0                 & 20.0                 \\
Overconfidence Bias                             & 20.0                                 & 80.0                                & 73.3                 & 26.7                 & 83.3                 & 16.7                 \\
Regression Fallacy                              & 20.0                                 & 80.0                                & 54.3                 & 45.7                 & 82.9                 & 17.1                 \\
Sunk Cost Fallacy                               & 0.0                                  & 100.0                               & 40.0                 & 60.0                 & 66.7                 & 33.3                 \\
Total                                           & 33.2                                 & 66.8                                & 68.3                 & 31.7                 & 79.0                 & 21.0                 \\ \Xhline{1.2px}
                                                & \multicolumn{1}{l}{}                 & \multicolumn{1}{l}{}                & \multicolumn{1}{l}{} & \multicolumn{1}{l}{} & \multicolumn{1}{l}{} & \multicolumn{1}{l}{} \\ \Xhline{1.2px}
\multicolumn{1}{c}{\multirow{2}{*}{Gemini 1.0 Pro}}     & \multicolumn{2}{c}{\begin{tabular}[c]{@{}c@{}}Standard\end{tabular}} & \multicolumn{2}{c}{GBI}                     & \multicolumn{2}{c}{SBI}                     \\ \cline{2-7} 
\multicolumn{1}{c}{}                            & Accuracy                             & Error Rate                          & Accuracy             & Error Rate           & Accuracy             & Error Rate           \\ \hline
Base Rate Fallacy                               & 10.0                                 & 90.0                                & 55.0                 & 45.0                 & 65.0                 & 35.0                 \\
Gambler's Fallacy                               & 90.0                                 & 10.0                                & 100.0                & 0.0                  & 100.0                & 0.0                  \\
Insensitivity to Sample Size                    & 90.0                                 & 10.0                                & 93.3                 & 6.7                  & 93.3                 & 6.7                  \\
Conjunction Fallacy                             & 26.7                                 & 73.3                                & 66.7                 & 33.3                 & 66.7                 & 33.3                 \\
Anchoring Bias                                  & 80.0                                 & 20.0                                & 65.0                 & 35.0                 & 80.0                 & 20.0                 \\
Overconfidence Bias                             & 40.0                                 & 60.0                                & 86.7                 & 13.3                 & 93.3                 & 6.7                  \\
Regression Fallacy                              & 25.7                                 & 74.3                                & 57.1                 & 42.9                 & 85.7                 & 14.3                 \\
Sunk Cost Fallacy                               & 6.7                                  & 93.3                                & 46.7                 & 53.3                 & 40.0                 & 60.0                 \\
Total                                           & 44.4                                 & 55.6                                & 71.2                 & 28.8                 & 80.0                 & 20.0                 \\ \Xhline{1.2px}
                                                & \multicolumn{1}{l}{}                 & \multicolumn{1}{l}{}                & \multicolumn{1}{l}{} & \multicolumn{1}{l}{} & \multicolumn{1}{l}{} & \multicolumn{1}{l}{} \\ \Xhline{1.2px}
\multicolumn{1}{c}{\multirow{2}{*}{LLaMA3-70B}} & \multicolumn{2}{c}{\begin{tabular}[c]{@{}c@{}}Standard\end{tabular}} & \multicolumn{2}{c}{GBI}                     & \multicolumn{2}{c}{SBI}                     \\ \cline{2-7} 
\multicolumn{1}{c}{}                            & Accuracy                             & Error Rate                          & Accuracy             & Error Rate           & Accuracy             & Error Rate           \\ \hline
Base Rate Fallacy                               & 15.0                                 & 85.0                                & 52.5                 & 47.5                 & 35.0                 & 65.0                 \\
Gambler's Fallacy                               & 20.0                                 & 80.0                                & 50.0                 & 50.0                 & 80.0                 & 20.0                 \\
Insensitivity to Sample Size                    & 50.0                                 & 50.0                                & 10.0                 & 90.0                 & 23.3                 & 76.7                 \\
Conjunction Fallacy                             & 13.3                                 & 86.7                                & 46.7                 & 53.3                 & 53.3                 & 46.7                 \\
Anchoring Bias                                  & 25.0                                 & 75.0                                & 15.0                 & 85.0                 & 65.0                 & 35.0                 \\
Overconfidence Bias                             & 33.3                                 & 66.7                                & 86.7                 & 13.3                 & 80.0                 & 20.0                 \\
Regression Fallacy                              & 5.7                                  & 94.3                                & 22.9                 & 77.1                 & 22.9                 & 77.1                 \\
Sunk Cost Fallacy                               & 13.3                                 & 86.7                                & 80.0                 & 20.0                 & 86.7                 & 13.3                 \\
Total                                           & 22.4                                 & 77.6                                & 43.9                 & 56.1                 & 50.2                 & 49.8                 \\ \Xhline{1.2px}
\end{tabular}}
\caption{Model GPT-4, Gemini 1.0 Pro and LLaMA3-70B prediction accuracy for Non-Abstention experiments (\%) on BRU dataset.}
\label{table3}
\end{table*}

\begin{table*}[]
\centering
\setlength{\tabcolsep}{1.7mm}{
\begin{tabular}{lcccccc}
\Xhline{1.5px}
\multicolumn{7}{c}{Abstention}                                                                                                                                                            \\ \Xhline{1.2px}
\multicolumn{1}{c}{\multirow{2}{*}{GPT-4}}      & \multicolumn{2}{c}{Standard}                & \multicolumn{2}{c}{GBI}                     & \multicolumn{2}{c}{SBI}                     \\ \cline{2-7} 
\multicolumn{1}{c}{}                            & Accuracy             & Error Rate           & Accuracy             & Error Rate           & Accuracy             & Error Rate           \\ \hline
Base Rate Fallacy (40)                               & 18.8                 & 65.0                 & 45.5                 & 15.0                 & 80.0                 & 7.5                  \\
Gambler's Fallacy (20)                               & 93.8                 & 5.0                  & 93.3                 & 5.0                  & 100.0                & 0.0                  \\
Insensitivity to Sample Size (30)                    & 96.6                 & 3.3                  & 92.6                 & 6.7                  & 100.0                & 0.0                  \\
Conjunction Fallacy (15)                             & 70.0                 & 20.0                 & N/A                  & N/A                  & 100.0                & 0.0                  \\
Anchoring Bias (20)                                  & 66.7                 & 25.0                 & 87.5                 & 5.0                  & 100.0                & 0.0                  \\
Overconfidence Bias (30)                             & 15.0                 & 56.7                 & N/A                  & N/A                  & 80.0                 & 6.7                  \\
Regression Fallacy (35)                              & 19.2                 & 60.0                 & 76.9                 & 8.6                  & 96.0                 & 2.9                  \\
Sunk Cost Fallacy (15)                               & 0.0                  & 33.3                 & 10.0                 & 60.0                 & 75.0                 & 13.3                 \\
Total (205)                                           & 48.4                 & 38.5                 & 73.8                 & 10.7                 & 93.5                 & 3.9                  \\ \Xhline{1.2px}
                                                & \multicolumn{1}{l}{} & \multicolumn{1}{l}{} & \multicolumn{1}{l}{} & \multicolumn{1}{l}{} & \multicolumn{1}{l}{} & \multicolumn{1}{l}{} \\ \Xhline{1.2px}
\multicolumn{1}{c}{\multirow{2}{*}{Gemini 1.0 Pro}}     & \multicolumn{2}{c}{Standard}                & \multicolumn{2}{c}{GBI}                     & \multicolumn{2}{c}{SBI}                     \\ \cline{2-7} 
\multicolumn{1}{c}{}                            & Accuracy             & Error Rate           & Accuracy             & Error Rate           & Accuracy             & Error Rate           \\ \hline
Base Rate Fallacy (40)                               & 15.8                 & 80.0                 & 46.7                 & 20.0                 & 72.7                 & 15.0                 \\
Gambler's Fallacy (20)                               & 100.0                & 0.0                  & 100.0                & 0.0                  & 100.0                & 0.0                  \\
Insensitivity to Sample Size (30)                    & 92.9                 & 6.7                  & 92.9                 & 6.7                  & 100.0                & 0.0                  \\
Conjunction Fallacy (15)                             & 35.7                 & 60.0                 & 50.0                 & 20.0                 & 93.3                 & 6.7                  \\
Anchoring Bias (20)                                  & 60.0                 & 30.0                 & 100.0                & 0.0                  & 100.0                & 0.0                  \\
Overconfidence Bias (30)                             & 36.8                 & 40.0                 & 100.0                & 0.0                  & 100.0                & 0.0                  \\
Regression Fallacy (35)                              & 28.1                 & 65.7                 & 81.3                 & 8.6                  & 96.4                 & 2.9                  \\
Sunk Cost Fallacy (15)                               & 0.0                  & 93.3                 & 16.7                 & 66.7                 & 44.4                 & 33.3                 \\
Total (205)                                           & 44.6                 & 47.8                 & 75.5                 & 12.7                 & 91.1                 & 6.3                  \\ \Xhline{1.2px}
                                                & \multicolumn{1}{l}{} & \multicolumn{1}{l}{} & \multicolumn{1}{l}{} & \multicolumn{1}{l}{} & \multicolumn{1}{l}{} & \multicolumn{1}{l}{} \\ \Xhline{1.2px}
\multicolumn{1}{c}{\multirow{2}{*}{LLaMA3-70B}} & \multicolumn{2}{c}{Standard}                & \multicolumn{2}{c}{GBI}                     & \multicolumn{2}{c}{SBI}                     \\ \cline{2-7} 
\multicolumn{1}{c}{}                            & Accuracy             & Error Rate           & Accuracy             & Error Rate           & Accuracy             & Error Rate           \\ \hline
Base Rate Fallacy (40)                               & 10.5                 & 85.0                 & 17.4                 & 47.5                 & 30.4                 & 40.0                 \\
Gambler's Fallacy (20)                               & 47.1                 & 45.0                 & 50.0                 & 30.0                 & 84.6                 & 10.0                 \\
Insensitivity to Sample Size (30)                    & 7.4                  & 83.3                 & 41.2                 & 33.3                 & 17.4                 & 63.3                 \\
Conjunction Fallacy (15)                             & 38.5                 & 53.3                 & 75.0                 & 6.7                  & 100.0                & 0.0                  \\
Anchoring Bias (20)                                  & 30.0                 & 70.0                 & 33.3                 & 50.0                 & 71.4                 & 20.0                 \\
Overconfidence Bias (30)                             & 13.3                 & 86.7                 & 75.9                 & 23.3                 & 89.7                 & 10.0                 \\
Regression Fallacy (35)                               & 5.7                  & 94.3                 & 3.7                  & 74.3                 & 16.7                 & 57.1                 \\
Sunk Cost Fallacy (15)                               & 36.4                 & 46.7                 & 20.0                 & 26.7                 & 91.7                 & 6.7                  \\
Total (205)                                           & 18.3                 & 76.1                 & 37.1                 & 40.5                 & 54.9                 & 31.7                 \\ \Xhline{1.2px}
\end{tabular}}
\caption{Model GPT-4, Gemini 1.0 Pro and LLaMA3-70B prediction accuracy for Abstention experiments (\%) on BRU dataset. 'N/A' indicates that there was no response in this category.}
\label{table4}
\end{table*}

\begin{table*}[]
\centering
\setlength{\tabcolsep}{5.2mm}{
\begin{tabular}{lccccc}
\Xhline{1.5px}
\multirow{2}{*}{Standard}    & \multicolumn{5}{c}{GPT-4}                                                                                        \\ \cline{2-6} 
                             & TT                   & TF                   & FT                   & FF                   & O                    \\ \hline
Base Rate Fallacy (40)            & 15.0                 & 0.0                  & 0.0                  & \textbf{65.0}        & 20.0                 \\
Gambler's Fallacy (20)            & \textbf{75.0}        & 0.0                  & 0.0                  & 5.0                  & 20.0                 \\
Insensitivity to Sample Size (30) & \textbf{93.3}        & 0.0                  & 0.0                  & 3.3                  & 3.3                  \\
Conjunction Fallacy (15)          & \textbf{46.7}        & 0.0                  & 0.0                  & 20.0                 & 33.3                 \\
Anchoring Bias (20)              & \textbf{50.0}        & 0.0                  & 0.0                  & 25.0                 & 25.0                 \\
Overconfidence Bias (30)          & 10.0                 & 0.0                  & 0.0                  & \textbf{56.7}        & 33.3                 \\
Regression Fallacy (35)           & 14.3                 & 0.0                  & 0.0                  & \textbf{60.0}        & 25.7                 \\
Sunk Cost Fallacy (15)            & 0.0                  & 0.0                  & 0.0                  & 33.3                 & \textbf{66.7}        \\
Total (205)                        & 36.1                 & 0.0                  & 0.0                  & \textbf{38.5}        & 25.4                 \\ \Xhline{1.5px}
                             & \multicolumn{1}{l}{} & \multicolumn{1}{l}{} & \multicolumn{1}{l}{} & \multicolumn{1}{l}{} & \multicolumn{1}{l}{} \\ \Xhline{1.5px}
\multirow{2}{*}{Standard}    & \multicolumn{5}{c}{Gemini 1.0 Pro}                                                                                       \\ \cline{2-6} 
                             & TT                   & TF                   & FT                   & FF                   & O                    \\ \hline
Base Rate Fallacy (40)            & 15.0                 & 0.0                  & 0.0                  & \textbf{80.0}        & 5.0                  \\
Gambler's Fallacy (20)            & \textbf{85.0}        & 0.0                  & 0.0                  & 0.0                  & 15.0                 \\
Insensitivity to Sample Size (30) & \textbf{86.7}        & 0.0                  & 0.0                  & 6.7                  & 6.7                  \\
Conjunction Fallacy (15)          & 33.3                 & 0.0                  & 0.0                  & \textbf{60.0}        & 6.7                  \\
Anchoring Bias (20)              & \textbf{45.0}        & 0.0                  & 0.0                  & 30.0                 & 25.0                 \\
Overconfidence Bias (30)          & 23.3                 & 0.0                  & 0.0                  & \textbf{40.0}        & 36.7                 \\
Regression Fallacy (35)           & 25.7                 & 0.0                  & 0.0                  & \textbf{65.7}        & 8.6                  \\
Sunk Cost Fallacy (15)            & 0.0                  & 0.0                  & 0.0                  & \textbf{93.3}        & 6.7                  \\
Total (205)                        & 38.5                 & 0.0                  & 0.0                  & \textbf{47.8}        & 13.7                 \\ \Xhline{1.5px}
                             & \multicolumn{1}{l}{} & \multicolumn{1}{l}{} & \multicolumn{1}{l}{} & \multicolumn{1}{l}{} & \multicolumn{1}{l}{} \\ \Xhline{1.5px}
\multirow{2}{*}{Standard}    & \multicolumn{5}{c}{LLaMA3-70B}                                                                                   \\ \cline{2-6} 
                             & TT                   & TF                   & FT                   & FF                   & O                    \\ \hline
Base Rate Fallacy (40)            & 10.0                 & 0.0                  & 0.0                  & \textbf{85.0}        & 5.0                  \\
Gambler's Fallacy (20)            & 40.0                 & 0.0                  & 0.0                  & \textbf{45.0}        & 15.0                 \\
Insensitivity to Sample Size (30) & 6.7                  & 0.0                  & 0.0                  & \textbf{83.3}        & 10.0                 \\
Conjunction Fallacy (15)          & 26.7                 & 0.0                  & 6.7                  & \textbf{53.3}        & 13.3                 \\
Anchoring Bias (20)              & 30.0                 & 0.0                  & 0.0                  & \textbf{70.0}        & 0.0                  \\
Overconfidence Bias (30)          & 13.3                 & 0.0                  & 0.0                  & \textbf{86.7}        & 0.0                  \\
Regression Fallacy (35)           & 5.7                  & 0.0                  & 0.0                  & \textbf{94.3}        & 0.0                  \\
Sunk Cost Fallacy (15)            & 26.7                 & 0.0                  & 0.0                  & \textbf{46.7}        & 26.7                 \\
Total (205)                        & 16.6                 & 0.0                  & 0.5                  & \textbf{76.1}        & 6.8                  \\ \Xhline{1.5px}
\end{tabular}}
\caption{The percentage (\%) of TT, TF, FT, FF, and O in abstention experiment for GPT-4, Gemini 1.0 Pro, and LLaMA3-70B on the BRU dataset using standard prompting strategies.}
\label{tab:table5}
\end{table*}

\begin{table*}[]
\centering
\setlength{\tabcolsep}{5.2mm}{
\begin{tabular}{lccccc}
\Xhline{1.5px}
\multirow{2}{*}{GBI}         & \multicolumn{5}{c}{GPT-4}                                                                                        \\ \cline{2-6} 
                             & TT                   & TF                   & FT                   & FF                   & O                    \\ \hline
Base Rate Fallacy (40)            & 12.5                 & 0.0                  & 0.0                  & 15.0                 & \textbf{72.5}        \\
Gambler's Fallacy (20)            & \textbf{70.0}        & 0.0                  & 0.0                  & 5.0                  & 25.0                 \\
Insensitivity to Sample Size (30) & \textbf{83.3}        & 0.0                  & 0.0                  & 6.7                  & 10.0                 \\
Conjunction Fallacy (15)          & 0.0                  & 0.0                  & 0.0                  & 0.0                  & \textbf{100.0}       \\
Anchoring Bias (20)               & 35.0                 & 0.0                  & 0.0                  & 5.0                  & \textbf{60.0}        \\
Overconfidence Bias (30)          & 0.0                  & 0.0                  & 0.0                  & 0.0                  & \textbf{100.0}       \\
Regression Fallacy (35)           & 28.6                 & 0.0                  & 0.0                  & 8.6                  & \textbf{62.9}        \\
Sunk Cost Fallacy (15)            & 6.7                  & 0.0                  & 0.0                  & \textbf{60.0}        & 33.3                 \\
Total (205)                        & 30.2                 & 0.0                  & 0.0                  & 10.7                 & \textbf{59.0}        \\ \Xhline{1.5px}
                             & \multicolumn{1}{l}{} & \multicolumn{1}{l}{} & \multicolumn{1}{l}{} & \multicolumn{1}{l}{} & \multicolumn{1}{l}{} \\ \Xhline{1.5px}
\multirow{2}{*}{GBI}         & \multicolumn{5}{c}{Gemini 1.0 Pro}                                                                                       \\ \cline{2-6} 
                             & TT                   & TF                   & FT                   & FF                   & O                    \\ \hline
Base Rate Fallacy (40)            & 17.5                 & 0.0                  & 0.0                  & 20.0                 & \textbf{62.5}        \\
Gambler's Fallacy (20)            & \textbf{85.0}        & 0.0                  & 0.0                  & 0.0                  & 15.0                 \\
Insensitivity to Sample Size (30) & \textbf{86.7}        & 0.0                  & 0.0                  & 6.7                  & 6.7                  \\
Conjunction Fallacy (15)          & 20.0                 & 0.0                  & 0.0                  & \textbf{20.0}        & \textbf{60.0}        \\
Anchoring Bias (20)               & 40.0                 & 0.0                  & 0.0                  & 0.0                  & \textbf{60.0}        \\
Overconfidence Bias (30)          & 13.3                 & 0.0                  & 0.0                  & 0.0                  & \textbf{86.7}        \\
Regression Fallacy (35)           & 37.1                 & 0.0                  & 0.0                  & 8.6                  & \textbf{54.3}        \\
Sunk Cost Fallacy (15)            & 13.3                 & 0.0                  & 0.0                  & \textbf{66.7}        & 20.0                 \\
Total (205)                         & 39.0                 & 0.0                  & 0.0                  & 12.7                 & \textbf{48.3}        \\ \Xhline{1.5px}
                             & \multicolumn{1}{l}{} & \multicolumn{1}{l}{} & \multicolumn{1}{l}{} & \multicolumn{1}{l}{} & \multicolumn{1}{l}{} \\ \Xhline{1.5px}
\multirow{2}{*}{GBI}         & \multicolumn{5}{c}{LLaMA3-70B}                                                                                   \\ \cline{2-6} 
                             & TT                   & TF                   & FT                   & FF                   & O                    \\ \hline
Base Rate Fallacy (40)            & 10.0                 & 0.0                  & 0.0                  & \textbf{47.5}        & 42.5                 \\
Gambler's Fallacy (20)            & 30.0                 & 0.0                  & 0.0                  & 30.0                 & \textbf{40.0}        \\
Insensitivity to Sample Size (30) & 23.3                 & 0.0                  & 0.0                  & 33.3                 & \textbf{43.3}        \\
Conjunction Fallacy (15)          & 20.0                 & 0.0                  & 0.0                  & 6.7                  & \textbf{73.3}        \\
Anchoring Bias (20)               & 25.0                 & 0.0                  & 0.0                  & \textbf{50.0}        & 25.0                 \\
Overconfidence Bias (30)          & \textbf{73.3}        & 0.0                  & 0.0                  & 23.3                 & 3.3                  \\
Regression Fallacy (35)           & 2.9                  & 0.0                  & 0.0                  & \textbf{74.3}        & 22.9                 \\
Sunk Cost Fallacy (15)            & 6.7                  & 0.0                  & 0.0                  & 26.7                 & \textbf{66.7}        \\
Total (205)                        & 23.9                 & 0.0                  & 0.0                  & \textbf{40.5}        & 35.6                 \\ \Xhline{1.5px}
\end{tabular}}
\caption{The percentage (\%) of TT, TF, FT, FF, and O in abstention experiment for GPT-4, Gemini 1.0 Pro, and LLaMA3-70B on the BRU dataset using GBI prompting strategies.}
\label{tab:table6}
\end{table*}

\begin{table*}[]
\centering
\setlength{\tabcolsep}{5.2mm}{
\begin{tabular}{lccccc}
\Xhline{1.5px}
\multirow{2}{*}{SBI}         & \multicolumn{5}{c}{GPT-4}                                                                                        \\ \cline{2-6} 
                             & TT                   & TF                   & FT                   & FF                   & O                    \\ \hline
Base Rate Fallacy (40)            & 30.0                 & 0.0                  & 0.0                  & 7.5                  & \textbf{62.5}        \\
Gambler's Fallacy (20)            & \textbf{90.0}        & 0.0                  & 0.0                  & 0.0                  & 10.0                 \\
Insensitivity to Sample Size (30) & \textbf{96.7}        & 0.0                  & 0.0                  & 0.0                  & 3.3                  \\
Conjunction Fallacy (15)          & 46.7                 & 0.0                  & 0.0                  & 0.0                  & \textbf{53.3}        \\
Anchoring Bias (20)               & \textbf{55.0}        & 0.0                  & 0.0                  & 0.0                  & 45.0                 \\
Overconfidence Bias (30)          & 26.7                 & 3.3                  & 0.0                  & 3.3                  & \textbf{66.7}        \\
Regression Fallacy (35)           & \textbf{68.6}        & 0.0                  & 0.0                  & 2.9                  & 28.6                 \\
Sunk Cost Fallacy (15)            & 40.0                 & 0.0                  & 0.0                  & 13.3                 & \textbf{46.7}        \\
Total (205)                        & \textbf{56.1}        & 0.5                  & 0.0                  & 3.4                  & 40.0                 \\ \Xhline{1.5px}
                             & \multicolumn{1}{l}{} & \multicolumn{1}{l}{} & \multicolumn{1}{l}{} & \multicolumn{1}{l}{} & \multicolumn{1}{l}{} \\ \Xhline{1.5px}
\multirow{2}{*}{SBI}         & \multicolumn{5}{c}{Gemini 1.0 Pro}                                                                                       \\ \cline{2-6} 
                             & TT                   & TF                   & FT                   & FF                   & O                    \\ \hline
Base Rate Fallacy (40)            & 40.0                 & 0.0                  & 0.0                  & 15.0                 & \textbf{45.0}        \\
Gambler's Fallacy (20)            & \textbf{95.0}        & 0.0                  & 0.0                  & 0.0                  & 5.0                  \\
Insensitivity to Sample Size (30) & \textbf{100.0}       & 0.0                  & 0.0                  & 0.0                  & 0.0                  \\
Conjunction Fallacy (15)          & \textbf{93.3}        & 0.0                  & 0.0                  & 6.7                  & 0.0                  \\
Anchoring Bias (20)               & 40.0                 & 0.0                  & 0.0                  & 0.0                  & \textbf{60.0}        \\
Overconfidence Bias (30)          & \textbf{50.0}        & 0.0                  & 0.0                  & 0.0                  & \textbf{50.0}        \\
Regression Fallacy (35)           & \textbf{77.1}        & 0.0                  & 0.0                  & 2.9                  & 20.0                 \\
Sunk Cost Fallacy (15)            & 26.7                 & 0.0                  & 0.0                  & 33.3                 & \textbf{40.0}        \\
Total (205)                        & \textbf{64.9}        & 0.0                  & 0.0                  & 6.3                  & 28.8                 \\ \Xhline{1.5px}
                             & \multicolumn{1}{l}{} & \multicolumn{1}{l}{} & \multicolumn{1}{l}{} & \multicolumn{1}{l}{} & \multicolumn{1}{l}{} \\ \Xhline{1.5px}
\multirow{2}{*}{SBI}         & \multicolumn{5}{c}{LLaMA3-70B}                                                                                   \\ \cline{2-6} 
                             & TT                   & TF                   & FT                   & FF                   & O                    \\ \hline
Base Rate Fallacy (40)            & 17.5                 & 0.0                  & 0.0                  & 40.0                 & \textbf{42.5}        \\
Gambler's Fallacy (20)            & \textbf{55.0}        & 0.0                  & 0.0                  & 10.0                 & 35.0                 \\
Insensitivity to Sample Size (30) & 13.3                 & 3.3                  & 0.0                  & \textbf{60.0}        & 23.3                 \\
Conjunction Fallacy (15)          & 33.3                 & 0.0                  & 6.7                  & 0.0                  & \textbf{60.0}        \\
Anchoring Bias (20)               & \textbf{50.0}        & 0.0                  & 0.0                  & 20.0                 & 30.0                 \\
Overconfidence Bias (30)          & \textbf{83.3}        & 0.0                  & 0.0                  & 13.3                 & 3.3                  \\
Regression Fallacy (35)           & 11.4                 & 0.0                  & 0.0                  & \textbf{57.1}        & 31.4                 \\
Sunk Cost Fallacy (15)            & \textbf{73.3}        & 0.0                  & 0.0                  & 6.7                  & 20.0                 \\
Total (205)                        & \textbf{37.6}        & 0.5                  & 0.5                  & 31.7                 & 29.8                 \\ \Xhline{1.5px}
\end{tabular}}
\caption{The percentage (\%) of TT, TF, FT, FF, and O in abstention experiment for GPT-4, Gemini 1.0 Pro, and LLaMA3-70B on the BRU dataset using SBI prompting strategies.}
\label{tab:table7}
\end{table*}

\begin{table*}[]
\centering
\begin{tabular}{lccc}
\Xhline{1.5px}
\multicolumn{1}{c}{\multirow{2}{*}{GPT-4o}} & \multicolumn{3}{c}{Transformation}                     \\ \cline{2-4} 
\multicolumn{1}{c}{}                        & Direct matching & Indirect matching & Overall matching \\ \hline
Anchoring Bias (20)                              & 65.0            & 0.0               & 65.0             \\
Base Rate Fallacy (40)                          & 45.0            & 42.5              & 87.5             \\
Conjunction Fallacy (15)                         & 13.3            & \textbf{86.7}              & \textbf{100.0}            \\
Gambler's Fallacy (20)                           & 95.0            & 0.0               & 95.0             \\
Insensitivity to Sample Size (30)                & 90.0            & 0.0               & 90.0             \\
Overconfidence Bias (30)                         & 83.3            & 0.0               & 83.3             \\
Regression Fallacy (35)                          & 45.7            & 0.0               & 45.7             \\
Sunk Cost Fallacy (15)                           & \textbf{100.0}           & 0.0               & \textbf{100.0}            \\
Total (205)                                       & 65.9            & 14.6              & 80.5             \\ \Xhline{1.5px}
\end{tabular}
\caption{This table presents the accuracy (\%) of GPT-4o in detecting specific bias traps within the Bias Detection Loop. "Direct matching" refers to instances where GPT-4o accurately identifies the exact subtype associated with a particular bias. In contrast, "indirect matching" denotes cases where GPT-4o recognizes either the broader parent category or a synonym of the bias subtype.}
\label{tab:table-trans}
\end{table*}

\begin{table*}[]
\centering
\begin{tabular}{>{\centering\arraybackslash}p{2cm}c>{\raggedright\arraybackslash}p{9cm}>{\raggedright\arraybackslash}p{3cm}}
\Xhline{1.5px}
\multicolumn{4}{c}{\textbf{Base Rate Fallacy}}                                                                                                                                                                                                                                                                                                                                                                                                                                                                                                                                                                                                                                    \\ \hline
Type                                                                  & Number               & \multicolumn{1}{c}{Question Example}        & Reference Source                                                                                                                                                                                                                                                                                                                                                                                                                                                                                                                                      \\ \hline
Questions Containing Numerical Information    & 15                   & There are two majors in a university: psychology and computer science. There are 700 students majoring in psychology and 300 students majoring in computer science. In an academic competition, a student won the best paper award in the school. This paper explores the content of artificial intelligence. Based on this information, which major do you think this student is most likely to come from? Option: A. Psychology major B. Computer Science Major  & Question prototype originally from \cite{tversky1974judgment}  \\ \Xhline{0.1px}
Questions Without Numerical Information & 25                   & There is a lady named Sarah who often buys lottery tickets and shows a strong interest in probability and statistical analysis. She often shares posts about math problems on social media and participates in various math games and challenges. Based on your understanding of Sarah, what profession do you think she is more likely to be? Option: A. He is a supermarket cashier B. He is a mathematics professor    & Question prototype originally from \cite{tversky1974judgment}     \\ \Xhline{1.5px}
\multicolumn{1}{l}{}                                                  & \multicolumn{1}{l}{} &                                                                                                                                                                                                                                                                                                                                                                                                                                                                                                                                                                           \\ \Xhline{1.5px}
\multicolumn{4}{c}{\textbf{Gambler's Fallacy}}      \\ \hline
Type                                                                  & Number               & \multicolumn{1}{c}{Question Example}   & Reference Source       \\ \hline
Questions Containing Numerical Information    & 10                   & In an online game, players can draw virtual cards, each with a different rarity level. Assuming that the probability of obtaining the rarest card in each draw is 1\%. A player has drawn 99 times but has not yet drawn any rare cards. Considering that the player has not won rare cards in the previous 99 draws, please choose which of the following options best describes the probability of the player winning rare cards in the 100th draw: Option: A. Above 1\% B. Less than 1\% C. Equal to 1\%  &    Question prototype originally from \cite{kahneman2011thinking} \\ \Xhline{0.1px}
Questions Without Numerical Information & 10                   & In a board game, a player fails to win any game in consecutive rounds. Considering that this player has failed several consecutive games, please choose which of the following options best describes his probability of winning the game in the next round: Option: A. Same B. Lower C. Higher    &    Question prototype originally from \cite{kahneman2011thinking}   \\ \Xhline{1.5px}
\end{tabular}
\caption{Examples of dataset questions. Here are Base Rate Fallacy and Gambler’s Fallacy, which include question types and corresponding quantities. The table categorizes questions into two types: "Questions Containing Numerical Information," where explicit numerical data is provided, and "Questions Without Numerical Information," where no specific numbers are included.}
\label{table8}
\end{table*}

\begin{table*}[]
\centering
\begin{tabular}{>{\centering\arraybackslash}p{2cm}c>{\raggedright\arraybackslash}p{9cm}>{\raggedright\arraybackslash}p{3cm}}
\Xhline{1.5px}
\multicolumn{4}{c}{\textbf{Anchoring Bias}}                                                                                                                                                         \\ \hline
Type                         & Number & \multicolumn{1}{c}{Question Example}      &  Reference Source    \\ \hline
Active Selection Questions   & 5      & Apple launched their Vision Pro glasses for \$3500 at the start of the year. Three months afterwards, a competitor introduced VR glasses that are lighter and \$2000 less expensive. Given that you seldom use VR glasses, which product would you consider buying? Option: A. Apple's Vision Pro glasses B. The lighter, less expensive VR glasses from a competitor C. None, as I rarely use VR glasses D. Explore other brands or wait for price reductions   &    Question prototype originally from \cite{kahneman2011thinking}           \\ \Xhline{0.1px}
Objective Analysis Questions & 15     & Before e-commerce became popular, the traditional economy flourished, and most consumers preferred shopping in brick-and-mortar stores. Despite instances of arbitrary price hikes in these stores, where many products were sold for much more than their worth, consumers continued to shop frequently. The rise of e-commerce brought some order, making pricing between physical stores and online platforms more transparent and competitive, leading to lower profit margins. Yet, even as the pricing chaos in physical stores was addressed, many consumers still chose to shop online instead of returning to physical stores. Which psychological effect is most likely driving this behavior? Option: A. Anchoring bias B. Conformity effect C. Loss aversion D. Confirmation bias  &    Question prototype originally from \cite{kahneman2011thinking}  \\ \Xhline{1.5px}
                             &        &             \\ \Xhline{1.5px}
\multicolumn{4}{c}{\textbf{Overconfidence Bias}}                                                                                                                                                                                                                                                                                                                                                                                                                                                                                                                                                                                                                                                                                                                                                                                               \\ \hline
Type                         & Number & \multicolumn{1}{c}{Question Example}  &  Reference Source                                                                                                                                                                                                                                                                                                                                                                                                                                                                                                                                                                                                                                                                                                                                                               \\ \hline
Objective Analysis Questions & 30     & Mr Li is preparing to tackle a tough professional certification exam in two weeks. He is highly confident in his exam readiness, convinced he's mastered all the required knowledge and skills. Additionally, he often discusses key topics and potential exam pitfalls with his friends and colleagues. What outcome is most probable in this scenario? Option: A. Mr. Li performs poorly. B. Mr. Li excels, surpassing expectations in his exam preparation and performance.         &    Question prototype originally from \cite{kahneman2011thinking}                                                                                                                                                                                                                                                                                                  \\ \Xhline{1.5px}
\end{tabular}
\caption{Examples of dataset questions. Here are Base Rate Fallacy and Gambler’s Fallacy, which include question types and corresponding quantities. The table categorizes questions into two types: "Active Selection Questions," which require participants to make an explicit choice between options based on a scenario, and "Objective Analysis Questions," which prompt participants to analyze a situation and identify the underlying cognitive bias or psychological effect.}
\label{table9}
\end{table*}

\begin{table*}[]
\centering
\begin{tabular}{>{\centering\arraybackslash}p{2cm}c>{\raggedright\arraybackslash}p{9cm}>{\raggedright\arraybackslash}p{3cm}}
\Xhline{1.5px}
\multicolumn{4}{c}{\textbf{Insensitivity to Sample Size}}   \\ \hline
\multicolumn{1}{c}{Type}               & Number               & \multicolumn{1}{c}{Question Example} &  Reference Source     \\ \hline
Sample Size Discrepancy of 10-fold     & 15                   & Assume two pet stores of varying sizes: one is a large chain and the other a small family-run business. Both stores track the monthly percentage of pet medications sold for treating illnesses. The large chain sells 1,000 bottles of pet medicine each month, whereas the small store sells about 100 bottles. Annually, each store typically sells 40\% of its medications for disease treatment. Which pet store is more likely to sell over 60\% of its medications for disease treatment in a given month? Option: A. The large chain pet store B. The small family-run pet shop
C. Both stores have similar sales ratios (with less than a 5\% difference) & Question prototype originally from \cite{tversky1974judgment}  \\ \Xhline{0.1px}
Sample Size Discrepancy of 10 to 100-fold & 15                   & Consider two bookstores: one a large chain and the other a small independent shop. Both track their monthly book sales, with the large chain selling 10,000 books and the small shop selling about 1,000 books per month. Typically, 30\% of each bookstore's monthly sales come from bestsellers. Which bookstore is more likely to have bestsellers make up over 60\% of its total sales in a given month? Option: A. The large chain bookstore B. The small independent bookstore C. The sales proportions of both bookstores are similar (with less than a 5\% difference)      & Question prototype originally from \cite{tversky1974judgment}                           \\ \Xhline{1.5px}
                                       & \multicolumn{1}{l}{} &                                                                                                                                                                                                                                                                                                                                                                                                                                                                                                                                                                                                                                                                                                                                                                                                                                   \\ \Xhline{1.5px}
\multicolumn{4}{c}{\textbf{Conjunction Fallacy}}                                                                                                                                                                                                                                                                                                                                                                                                                                                                                                                                                                                                                                                                                                                                                                                                                                                           \\ \hline
\multicolumn{1}{c}{Type}               & Number               & \multicolumn{1}{c}{Question Example} &  Reference Source      \\ \hline
A $\to$ B Paradigms                          & 15                   & Which of the following is more common?  A. An eight-letter word, the sixth letter is "e"  B. Eight-letter words ending in "- ness".       & Question prototype originally from \cite{Conj}                                                                                                                                                                                                                                                                                                                                                                                                                                                                                                                                                                                                                                                              \\ \Xhline{1.5px}
\end{tabular}
\caption{Examples of dataset questions. Here are Insensitivity to Sample Size and Conjunction Fallacy, which include question types and corresponding quantities. The table categorizes Insensitivity to Sample Size questions into two types: "Sample Size Discrepancy of 10-fold," where the comparison is made between two groups with a tenfold difference in sample size, and "Sample Size Discrepancy of 10 to 100-fold," where the sample size difference is broader, ranging from 10 to 100 times.}
\label{table10}
\end{table*}

\begin{table*}[]
\centering
\begin{tabular}{>{\centering\arraybackslash}p{2cm}c>{\raggedright\arraybackslash}p{9cm}>{\raggedright\arraybackslash}p{3cm}}
\Xhline{1.5px}
\multicolumn{4}{c}{\textbf{Regression Fallacy}}   \\ \hline
Type                            & Number               & \multicolumn{1}{c}{Question Example}   & Reference Source     \\ \hline
Single-Event Assessment Questions    & 20                   & Mr Wang, an accountant, discovered during a routine health check that his cholesterol levels were higher than before, which concerned him due to a family history of heart disease. To address this, he adjusted his diet, reduced his intake of high-fat foods, and began working out at the gym. After three days, his cholesterol levels had notably decreased and returned to a healthy range. Which of the following is the most likely explanation for this reduction in cholesterol levels? Option: A. Changing dietary habits and regular exercise B. Using more effective medication for treatment C. Cholesterol levels naturally fluctuating back to normal D. Improved overall lifestyle and reduced work stress    & Question prototype originally from \cite{Regress}                                                                                                                                                                                                \\ \Xhline{0.1px}
Multiple-Event Assessment Questions & 15                   & Artist Ms Li's works were highly acclaimed and awarded at an international art exhibition, marking a significant milestone in her career. This acclaim boosted her popularity, leading to more exhibition opportunities and collaboration offers. To sustain her success, Ms. Li devoted more time and effort to creating new works, incorporating new artistic styles and techniques. Despite receiving positive critiques, her subsequent works did not achieve the same level of recognition and awards as her earlier pieces. Which of the following is the most likely reason Ms. Li's recent works did not replicate her previous success? Option: A. Her new artistic style did not resonate with her existing audience. B. New competitors emerged in the art market. C. There was a shift in the prevailing aesthetic trends in the art world. D. Her performance naturally returned to her regular performance level & Question prototype originally from \cite{Regress} \\ \Xhline{1.5px}    & \multicolumn{1}{l}{} &        \\ \Xhline{1.5px}
\multicolumn{4}{c}{\textbf{Sunk Cost Fallacy}}       \\ \hline
Type                            & Number               & \multicolumn{1}{c}{Question Example}  & Reference Source   \\ \hline
Objective Analysis Questions    & 15                   & A company has invested \$8 million in a technology project that was originally forecast to yield \$15 million in profits. However, due to changing market conditions, a reevaluation indicates that the expected profit upon completion would now only be \$7 million. The company must now consider whether to invest an additional \$4 could million to finish the project. What decision should the company make? Option: A. Proceed with the additional \$4 million investment to complete the project. B. Cease further investment and abandon the project. C. Explore options to reduce the completion costs. D. Reevaluate the project’s market potential to see if strategic adjustments could enhance the anticipated profit.      & Question prototype originally from \cite{Sunkcost}          \\ \Xhline{1.5px}
\end{tabular}
\caption{Examples of dataset questions. Here are Regression Fallacy and Sunk Cost Fallacy, which include question types and corresponding quantities. The table categorizes Regression Fallacy questions into two types: "Single-Event Assessment Questions," where participants assess a scenario based on a single event or change, and "Multiple-Event Assessment Questions," where the scenario involves a series of events or observations over time.}
\label{table11}
\end{table*}

\begin{table*}[]
\centering
\begin{tabular}{>{\centering\arraybackslash}p{2cm}>{\raggedright\arraybackslash}p{11cm}}
\Xhline{1.5px}
\multicolumn{1}{c}{\textbf{Field}}  & \multicolumn{1}{c}{\textbf{Description}}                                 \\ \hline
Query                      & \textbf{If you prefer not to make a decisive choice, then select option E.} In a large technology company, 35\% of the employees are engineers and 65\% are salespeople. The company recently held a technology innovation competition, and you know an employee who won the competition. Based on this information, which department do you think this employee is most likely from? Option: A. Engineering Department B. Sales Department E: I am not sure which choice is the best to select.                                                                                                                                                                                  \\ \Xhline{0.1px}
Ground Truth               & B. Sales Department                                                              \\ \Xhline{0.1px}
Model Reasoning & Based on the information given, it is more likely that the employee who won the technology innovation competition is from the Engineering Department. Here's why: The nature of the event, a technology innovation competition, suggests that participants with technical expertise and skills relevant to engineering would have an advantage and be more likely to succeed. Although there are more salespeople in the company, the specific context of a technology innovation competition leans towards the strengths and job functions typically found in the engineering sector. Therefore, the best guess would be: Option A: Engineering Department. \\ \Xhline{0.1px}
Model Answer & A: Engineering Department  \\ \Xhline{1.5px}
\end{tabular}
\caption{Examples of GPT-4 \textbf{Standard} prompt reasoning in abstention experiment about Base Rate Fallacy.}
\label{table12}
\end{table*}

\begin{table*}[]
\centering
\begin{tabular}{>{\centering\arraybackslash}p{2cm}>{\raggedright\arraybackslash}p{11cm}}
\Xhline{1.5px}
\multicolumn{1}{c}{\textbf{Field}}  & \multicolumn{1}{c}{\textbf{Description}}                                 \\ \hline
Query                      & \textbf{Please provide a definition of cognitive bias and identify any instances of these biases in the decision-making process. If you prefer not to make a decisive choice, then select option E.}
In a large technology company, 35\% of the employees are engineers and 65\% are salespeople. The company recently held a technology innovation competition, and you know an employee who won the competition. Based on this information, which department do you think this employee is most likely from? Option: A. Engineering Department B. Sales Department
E: I am not sure which choice is the best to select.                                                                                                                                                                               \\ \Xhline{0.1px}
Ground Truth               & B. Sales Department                                                             \\ \Xhline{0.1px}
Model Reasoning & Cognitive bias refers to systematic errors in thinking that affect decisions and judgments that people make. These biases often arise from the brain's attempt to simplify information processing and are influenced by an individual's personal experiences, preferences, or emotions.
Regarding the question about the employee's department:
Given the statistics provided—35\% of employees are engineers and 65\% are salespeople—and knowing that the competition is focused on technology innovation, it might be tempting to assume that the engineer has a higher likelihood of winning due to their likely closer engagement with technology-related tasks. However, without knowing the criteria of the competition, the specific roles of salespeople and engineers in the company, or the talents and capabilities of the individuals involved, it's not possible to definitively determine which department the winner is from based purely on the information given.
Therefore, in accordance with the instruction to avoid biases and not make a decisive choice without sufficient information, I would choose:
E. I am not sure which option is the best to select. \\ \Xhline{0.1px}
Model Answer & E. I am not sure which option is the best to select.  \\ \Xhline{1.5px}
\end{tabular}
\caption{Examples of GPT-4 \textbf{GBI} prompt reasoning in abstention experiment about Base Rate Fallacy.}
\label{table13}
\end{table*}

\begin{table*}[]
\centering
\begin{tabular}{>{\centering\arraybackslash}p{2cm}>{\raggedright\arraybackslash}p{11cm}}
\Xhline{1.5px}
\multicolumn{1}{c}{\textbf{Field}}  & \multicolumn{1}{c}{\textbf{Description}}                                 \\ \hline
Query                      & \textbf{Please provide a definition of the Base Rate Fallacy, then identify any instances of this specific bias in the decision-making process.  If you prefer not to make a decisive choice, then select option E.}
In a large technology company, 35\% of employees are engineers and 65\% are salespeople. The company recently held a technology innovation competition, and you know an employee who won the competition. Based on this information, which department do you think this employee is most likely from? Option: A. Engineering Department B. Sales Department
E: I am not sure which choice is the best to select.                                                                                                                                                                               \\ \Xhline{0.1px}
Ground Truth               & B. Sales Department                                                             \\ \Xhline{0.1px}
Model Reasoning & The Base Rate Fallacy refers to a common cognitive error where people neglect or underestimate the importance of the base rate (or general prevalence) information when making decisions under uncertainty. In essence, when given specific information, individuals tend to focus on that information and ignore the broader context that the base rates provide.
In the scenario you provided, the base rates are that 35\% of employees are engineers and 65\% are salespeople. If we have no other information about the likelihood of employees from either department winning the competition based on their job roles, then statistically, a randomly chosen winner is more likely to come from the sales department simply because there are more salespeople than engineers. However, if there are factors that could influence the likelihood of winning based on department, such as the nature of the competition favoring technical skills typically stronger in engineers, this would shift the probability. Without such additional information, the answer, strictly based on the provided base rates, would be: B. Sales Department. This follows from the fact that a larger proportion of the company's employees are salespeople. Thus, absent any other distinguishing information about the likelihood of winning, a winner chosen at random is more likely to be from the Sales Department, just due to sheer numbers. \\ \Xhline{0.1px}
Model Answer & B. Sales Department \\ \Xhline{1.5px}
\end{tabular}
\caption{Examples of GPT-4 \textbf{SBI} prompt reasoning in abstention experiment about Base Rate Fallacy.}
\label{table14}
\end{table*}

\begin{table*}[]
\centering
\begin{tabular}{>{\centering\arraybackslash}p{2cm}>{\raggedright\arraybackslash}p{11cm}}
\Xhline{1.5px}
\textbf{Field}            & \multicolumn{1}{c}{\textbf{Description}}     \\ \hline
Bias Detection prompting     & There are two majors in a university: psychology and computer science. There are 700 students majoring in psychology and 300 students majoring in computer science. In an academic competition, a student won the best paper award in the school. This paper explores the content of artificial intelligence. Based on this information, which major do you think this student is most likely to come from? Option: A. Psychology major B. Computer Science Major. \textbf{Please identify which cognitive bias trap is contained in this question and return the cognitive bias type. The most likely cognitive bias trap is}                  \\ \hline
Type          & Base Rate Fallacy        \\ \hline Model Reasoning & The most likely cognitive bias trap in this question is the "\textbf{base rate fallacy}" or "\textbf{base rate neglect}." This occurs when people ignore the base rate (general prevalence) of an event or characteristic in favor of specific information. In this scenario, the base rates are the number of students in each major: 700 psychology students and 300 computer science students. The specific information is that the paper explores artificial intelligence, which is more closely related to computer science. People might be inclined to choose the computer science major as the more likely source of the student who won the award because of the content of the paper, even though there are more psychology students overall. The base rate fallacy leads people to overlook the fact that, statistically, there are more psychology students, which could mean that a psychology student is more likely to have won simply due to their greater numbers, unless the specific content of the paper strongly indicates otherwise. \\ \hline Model Answer  & Base Rate Fallacy       \\ \Xhline{1.5px}
\end{tabular}
\caption{Examples of GPT-4o's reasoning in the Bias Detection Module regarding the Base Rate Fallacy. This instance demonstrates a direct match for the bias type within the given question.}
\label{table15}
\end{table*}

\begin{table*}[]
\centering
\begin{tabular}{>{\centering\arraybackslash}p{2cm}>{\raggedright\arraybackslash}p{11cm}}
\Xhline{1.5px}
\textbf{Field}        & \multicolumn{1}{c}{\textbf{Description}}         \\ \hline
Bias Detection prompting     & Which is more common in German A. Six letter words ending in "- ung" B. A six letter word, the fifth letter is "n"? \textbf{Please identify which cognitive bias trap is contained in this question and return the cognitive bias type. The most likely cognitive bias trap is}    \\ \hline
Type     & Conjunction Fallacy       \\ \hline
Model Reasoning & The question you're dealing with involves evaluating the likelihood of two different linguistic patterns in the German language. The cognitive bias that is most likely to occur here is the "\textbf{availability heuristic}".The availability heuristic is a mental shortcut that relies on immediate examples that come to a person's mind when evaluating a specific topic, concept, method, or decision. When people use the availability heuristic, they tend to overestimate the likelihood of events based on their ability to recall examples. In this case, people might more easily recall German words ending in "-ung" because "-ung" is a common suffix in German that forms nouns from verbs (similar to "-ing" or "-tion" in English). This might make people think that there are more six-letter words ending in "-ung" than there actually are, simply because the pattern is familiar and easy to recall. On the other hand, the pattern of a six-letter word with the fifth letter "n" is less distinctive and harder to recall specific examples for, even though it might actually be more common due to the flexibility of word formation and the frequency of the letter "n" in various positions. Thus, the availability heuristic might lead people to incorrectly judge the frequency of these patterns based on how easily examples come to mind, rather than on actual linguistic data. \\ \hline
Model Answer    & Availability Heuristic (parent category of Conjunction Fallacy)       \\ \Xhline{1.5px}
\end{tabular}
\caption{Examples of GPT-4o's reasoning in the Bias Detection Module regarding the Conjunction Fallacy. This instance demonstrates an indirect match for the parent category of bias type within the given question.}
\label{table16}
\end{table*}

\begin{table*}[]
\centering
\begin{tabular}{>{\centering\arraybackslash}p{2cm}>{\raggedright\arraybackslash}p{13.5cm}}
\Xhline{1.5px}
\textbf{Field}     & \multicolumn{1}{c}{\textbf{Overall Workflow Demonstration 1.1}}   \\ \hline
Abstention + GBI prompting  & \textcolor{gbi-color}{Please provide a definition of \textbf{cognitive bias} and identify any instances of these biases in the decision-making process. If you prefer not to make a decisive choice, then select option E.} In a long-distance race, a runner failed to take the lead in the past few races. The observer noticed the runner's consecutive failures. Considering the runner's performance in several consecutive races, please choose which of the following options best describes his probability of leading in the next race: Option: A. Higher B. Lower C. Same E. I am not sure which option is the best to select  \\ \hline Ground Truth   & C. Same       \\ \hline
Type     & Gambler's Fallacy         \\ \hline
Model Reasoning & Cognitive Bias Definition: A cognitive bias is a systematic pattern of deviation from norm or rationality in judgment. Individuals create their own "subjective reality" from their perception of the input. An individual's construction of reality, not the objective input, may dictate their behavior in the world. Thus, cognitive biases may sometimes lead to perceptual distortion, inaccurate judgment, illogical interpretation, or what is broadly called irrationality. Application to the Scenario: Given the description of cognitive bias and the emphasis on avoiding biases in decision-making, the most appropriate response to the question regarding the runner's probability of leading in the next race would be Option E: I am not sure which option is the best to select. This option avoids making a decision based on the observed sequence of events (his past consecutive failures), as this could involve the gambler's fallacy (believing that independent events in a sequence affect each other) or the availability heuristic (focusing on recent or easily remembered examples). Selecting "I am not sure" is a neutral position that acknowledges the limitations in predicting future outcomes based on past performances, especially when not enough information about other influencing factors (like changes in training, health, or competition) is available. \\ \hline
Model Answer   & \textcolor{decision-color}{E. I am not sure which option is the best to select.}    \\ \hline
Bias Detection prompting      & In a long-distance race, a runner failed to take the lead in the past few races. The observer noticed the runner's consecutive failures. Considering the runner's performance in several consecutive races, please choose which of the following options best describes his probability of leading in the next race: Option: A. Higher B. Lower C. Same \textcolor{detection-color}{Please identify which cognitive bias trap is contained in this question and return the cognitive bias type. The most likely cognitive bias trap is  }        \\ \hline Model Reasoning & The most likely cognitive bias trap in this situation is the "\textbf{Gambler's Fallacy}." This cognitive bias occurs when someone believes that past independent events can influence the likelihood of future independent events. In this context, each race is an independent event, and the runner's previous failures to take the lead do not affect the probability of leading in the next race.  \\ \hline
Model Answer   & \textcolor{decision-color}{Gambler's Fallacy}     \\ \Xhline{1.5px}
\end{tabular}
\caption{This table provides a detailed demonstration of the process depicted in Fig.~\ref{fig:fig2}. This is part 1, part 2 shows as follows. \textcolor{gbi-color}{Light green} represents the GBI prompt, \textcolor{sbi-color}{orange} indicates the SBI prompt, \textcolor{detection-color}{light blue} denotes the bias detection prompt, and \textcolor{decision-color}{purple} signifies the answer. The example illustrates how, upon the Bias Detection module \textbf{identifying the corresponding category} of the question type, the SBI prompt is triggered, successfully leading to the correct result.}
\label{table-demo1-1}
\end{table*}

\begin{table*}[]
\centering
\begin{tabular}{>{\centering\arraybackslash}p{3cm}>{\raggedright\arraybackslash}p{11cm}}
\Xhline{1.5px}
\textbf{Field}                & \multicolumn{1}{c}{\textbf{Overall Workflow Demonstration 1.2}}     \\ \hline
Abstention + SBI prompting    & \textcolor{sbi-color}{Please provide a definition of the Gambler's Fallacy, then identify any instances of this specific bias in the decision-making process.} In a long-distance race, a runner failed to take the lead in the past few races. The observer noticed the runner's consecutive failures. Considering the runner's performance in several consecutive races, please choose which of the following options best describes his probability of leading in the next race: Option: A. Higher B. Lower C. Same E. I am not sure which option is the best to select \\ \hline Ground Truth   & C. Same       \\ \hline
Type     & Gambler's Fallacy         \\ \hline
Model Reasoning & The Gambler's Fallacy is the mistaken belief that if something happens more frequently than normal during a given period, it will happen less frequently in the future, or vice versa. In other words, it is the belief that there are "due" occurrences that need to balance out the frequency of events, even when the events are independent of each other. Considering the Gambler's Fallacy and the independence of each race outcome in your scenario, the correct answer is: C. Same Each race is an independent event, and the runner’s probability of leading in the next race is the same as in any other race, assuming no changes in conditions or the runner's fitness. Therefore, their past performance does not influence their chances in future races unless there are relevant changes in variables that affect race outcomes. \\ \hline
Model Answer    &  \textcolor{decision-color}{C. Same}   \\ \Xhline{1.5px}
\end{tabular}
\caption{This table provides a detailed demonstration of the process depicted in Fig.~\ref{fig:fig2}. This is part 2. \textcolor{gbi-color}{Light green} represents the GBI prompt, \textcolor{sbi-color}{orange} indicates the SBI prompt, \textcolor{detection-color}{light blue} denotes the bias detection prompt, and \textcolor{decision-color}{purple} signifies the answer. The example illustrates how, upon the Bias Detection module \textbf{identifying the corresponding category} of the question type, the SBI prompt is triggered, successfully leading to the correct result.}
\label{table-demo1-2}
\end{table*}

\begin{table*}[]
\centering
\begin{tabular}{>{\centering\arraybackslash}p{2cm}>{\raggedright\arraybackslash}p{13.5cm}}
\Xhline{1.5px}
\textbf{Field}      & \multicolumn{1}{c}{\textbf{Overall Workflow Demonstration 2.1}}     \\ \hline
Abstention + GBI prompting                                  & \textcolor{gbi-color}{Please provide a definition of \textbf{cognitive bias} and identify any instances of these biases in the decision-making process. If you prefer not to make a decisive choice, then select option E.} In a large technology company, 35\% of employees are engineers and 65\% are salespeople. The company recently held a technology innovation competition, and you know an employee who won the competition. Based on this information, which department do you think this employee is most likely from? Option: A. Engineering Department B. Sales Department E. I am not sure which option is the best to select \\ \hline
Ground Truth    & B. Sales Department     \\ \hline
Type   & Base Rate Fallacy                                                                \\ \hline
Model Reasoning & Cognitive bias refers to systematic errors in thinking that affect decisions and judgments that people make. These biases often arise from the brain's attempt to simplify information processing and are influenced by an individual's personal experiences, preferences, or emotions. Regarding the question about the employee's department: Given the statistics provided—35\% of employees are engineers and 65\% are salespeople—and knowing that the competition is focused on technology innovation, it might be tempting to assume that the engineer has a higher likelihood of winning due to their likely closer engagement with technology-related tasks. However, without knowing the criteria of the competition, the specific roles of salespeople and engineers in the company, or the talents and capabilities of the individuals involved, it's not possible to definitively determine which department the winner is from based purely on the information given. Therefore, in accordance with the instruction to avoid biases and not make a decisive choice without sufficient information, I would choose: E. I am not sure which option is the best to select.   \\ \hline
Model Answer   & \textcolor{decision-color}{E. I am not sure which option is the best to select.}   \\ \hline
Bias Detection prompting   & In a large technology company, 35\% of employees are engineers and 65\% are salespeople. The company recently held a technology innovation competition, and you know an employee who won the competition. Based on this information, which department do you think this employee is most likely from? Option: A. Engineering Department B. Sales Department \textcolor{detection-color}{Please identify which cognitive bias trap is contained in this question and return the cognitive bias type. The most likely cognitive bias trap is}   \\ \hline
GPT-4o Model Reasoning & The most likely cognitive bias trap in this situation is the "\textbf{representativeness heuristic}." This cognitive bias occurs when people judge the probability or frequency of an event by how much it resembles a typical case or stereotype, rather than using actual statistical evidence. In this scenario, people might assume that because the competition is a "technology innovation competition," the winner is more likely to be an engineer, as engineers are stereotypically associated with technology and innovation. However, this ignores the actual \textbf{base rates} provided: 35\% of employees are engineers, and 65\% are salespeople. Without additional information about the likelihood of engineers versus salespeople winning the competition, the base rate suggests that a randomly selected employee is more likely to be from the sales department simply because there are more salespeople. Therefore, the representativeness heuristic \textbf{might lead someone to incorrectly assume} the winner is from the engineering department, even though statistically, there are more salespeople in the company.  \\ \hline GPT-4o Model Answer   & \textcolor{decision-color}{Representativeness Heuristic \ (parent category of Base Rate Fallacy)}   \\ \Xhline{1.5px}
\end{tabular}
\caption{This table provides a detailed demonstration of the process depicted in Fig.~\ref{fig:fig2}, with the colors in the table corresponding to those in the flowchart. This is part 1, part 2 shows as follows.}
\label{table-demo2-1}
\end{table*}

\begin{table*}[]
\centering
\begin{tabular}{>{\centering\arraybackslash}p{3cm}>{\raggedright\arraybackslash}p{11cm}}
\Xhline{1.5px}
\textbf{Field}       & \multicolumn{1}{c}{\textbf{Overall Workflow Demonstration 2.2}}     \\ \hline
Abstention + SBI prompting                                  & \textcolor{sbi-color}{Please provide a definition of the \textbf{representativeness heuristic}, then identify any instances of this specific bias in the decision-making process. If you prefer not to make a decisive choice, then select option E.} In a large technology company, 35\% of employees are engineers and 65\% are salespeople. The company recently held a technology innovation competition, and you know an employee who won the competition. Based on this information, which department do you think this employee is most likely from? Option: A. Engineering Department B. Sales Department E. I am not sure which option is the best to select                                                                                                                                                                                                                                                                                                                                                                                                                                                                                                                                              \\ \hline Ground Truth    & B. Sales Department     \\ \hline
Type   & Base Rate Fallacy                                                                \\ \hline
Model Reasoning & The representativeness heuristic is a mental shortcut that people use when making judgments about the probability of an event by comparing it to an existing prototype or stereotype in their minds. This heuristic leads people to overlook base rates (the actual statistical likelihood of something) and instead make decisions based on how representative or typical something seems. In your example, if someone uses the representativeness heuristic, they might assume the employee who won the technology innovation competition is more likely to be from the Engineering Department because engineering aligns more closely with technology innovation. This assumption, however, ignores the actual base rates: 35\% of employees are engineers, while 65\% are salespeople. To avoid this bias, you should consider the base rates. Since salespeople make up a larger proportion of the company (65\%), it is statistically more likely that the employee is from the Sales Department, despite the competition's technological focus. Therefore, the most likely correct choice would be: B. Sales Department \\ \hline
Model Answer   & \textcolor{decision-color}{B. Sales Department} \\ \Xhline{1.5px}
\end{tabular}
\caption{This table provides a detailed demonstration of the process depicted in Fig.~\ref{fig:fig2}. This is part 2. \textcolor{gbi-color}{Light green} represents the GBI prompt, \textcolor{sbi-color}{orange} indicates the SBI prompt, \textcolor{detection-color}{light blue} denotes the bias detection prompt, and \textcolor{decision-color}{purple} signifies the answer. The example illustrates how, upon the Bias Detection module \textbf{identifying the parent category} of the question type, the SBI prompt is triggered, successfully leading to the correct result. }
\label{table-demo2-2}
\end{table*}

\end{document}